%% file: paper_arxiv.tex
\documentclass{article} 

\input{math_commands.tex}

\usepackage{hyperref}
\usepackage{natbib}
\usepackage{url}
\usepackage{amsmath,amssymb,amsthm}
\usepackage{bm}
\usepackage{booktabs}
\usepackage{graphicx}
\usepackage{microtype}
\usepackage{enumitem}
\usepackage{xcolor}
\usepackage{multirow}
\usepackage{xr}
\usepackage{cleveref}

\newtheorem{theorem}{Theorem}[section]
\newtheorem{proposition}[theorem]{Proposition}
\newtheorem{lemma}[theorem]{Lemma}
\newtheorem{corollary}[theorem]{Corollary}
\newtheorem{definition}[theorem]{Definition}

\theoremstyle{remark}

\newcommand{\M}{\mathbb{M}}
\newcommand{\RR}{\mathbb{R}}
\newcommand{\Sphere}{\mathbb{S}^{2}}
\newcommand{\dcc}{d_{\mathrm{CC}}}

\newcommand{\SE}{\mathrm{SE}(3)}
\newcommand{\SOtwo}{\mathrm{SO}(2)}

\newcommand{\inner}[2]{\langle #1, #2 \rangle}
\usepackage{todonotes}

\title{Sub-Riemannian structure for Oriented Point Clouds}

\author{
    Thomas Besnier \qquad Stefan Sommer \\[0.5em]
    \small University of Copenhagen, Datalogisk Institute (DIKU)
}

\date{}

\begin{document}

\input{main}

\input{supplementary}

\bibliography{references}
\bibliographystyle{plain}

\end{document}

%% file: math_commands.tex
\usepackage{amsmath,amsfonts,bm}

\def\eqref#1{equation~\ref{#1}}

\def\1{\bm{1}}

\DeclareMathAlphabet{\mathsfit}{\encodingdefault}{\sfdefault}{m}{sl}
\SetMathAlphabet{\mathsfit}{bold}{\encodingdefault}{\sfdefault}{bx}{n}

\newcommand{\R}{\mathbb{R}}



%% file: main.tex
\maketitle

\begin{figure}[ht!]
    \centering
    \includegraphics[width=0.95\linewidth]{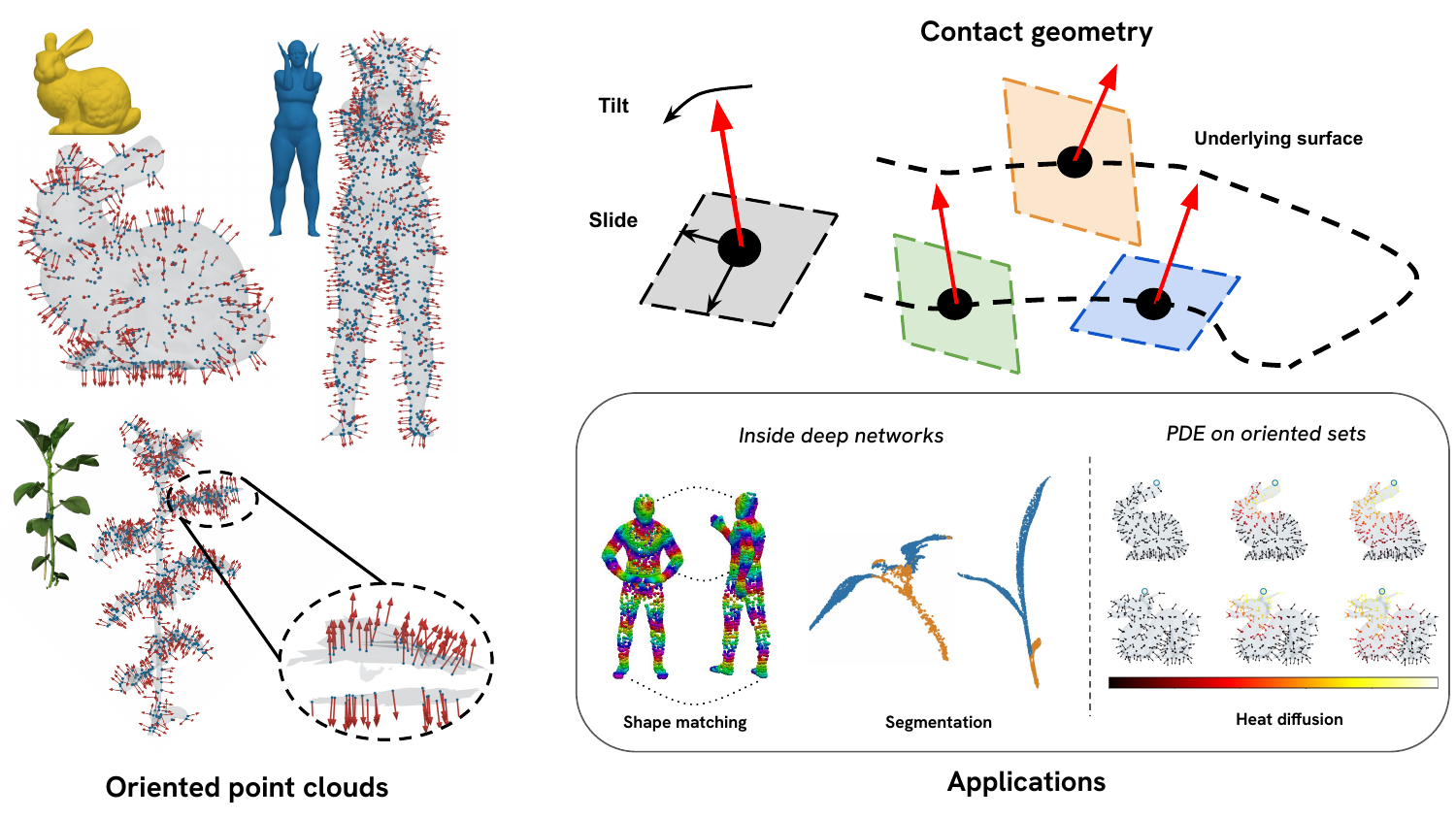}
    \vspace{-2mm}
    \caption{\textbf{We introduce a metric structure to define locality within oriented point clouds.} These are lifted to position–orientation space in which tangential translation and normal reorientation define a contact sub-Riemannian structure. An explicit pairwise score enables to compute local neighborhoods that can be used downstream for learning tasks and other shape processing applications such as heat diffusion on oriented point clouds.}
    \label{fig:thumbnail}
\end{figure}

\begin{abstract}
Many frameworks dealing with 3D (oriented) point clouds define locality in the ambient Euclidean space: $k$-nearest-neighbor graphs and radius balls in $\RR^3$ are critical to define which points are "close to one another" and should exchange information. This can unintentionally merge sampled surfaces that are close in space but geometrically distinct such as the two faces of a thin leaf. Then, features are more prone to leak across boundaries which hinder performance. To address this, we propose to define a sub-Riemannian contact metric on position-orientation space $\M=\RR^3\times \Sphere$: points are lifted to $(x,n)$ with $n$ a surface normal, and "motion" is constrained to a non-holonomic horizontal distribution in which spatial velocity must stay in the plane orthogonal to the normal which can reorient at a cost. From this, we derive an explicit score, which locally approximate the distance, to compare oriented points for downstream tasks. We report improved segmentation and correspondence accuracy near contact when using the score as a drop-in change inside PointNet++ aggregation mechanism as well as reduced leakage when using it to simulate heat diffusion on oriented points.
\end{abstract}

\section{Introduction}
\label{sec:intro}
To deal with unstructured 3D data, such as (oriented) point clouds, triangle soups and Gaussian splats \citep{kerbl3Dgaussians}, many networks \citep{qi2017pointnetpp,wang2019dynamic}, equivariant architectures \citep{thomas2018tensor,fuchs2020se3}, and classical geometry processing algorithms \citep{taubin1995signal,fleishman2003bilateral}, define locality as Euclidean. Neighborhoods are $k$-Nearest Neighbor ($k$NN) sets or radius balls in $\RR^3$, and convolutions, message passing, or diffusions act over them.

\begin{figure}[ht!]
    \centering
    \includegraphics[trim={0 2.5cm 0 2.4cm},clip,width=0.85\linewidth]{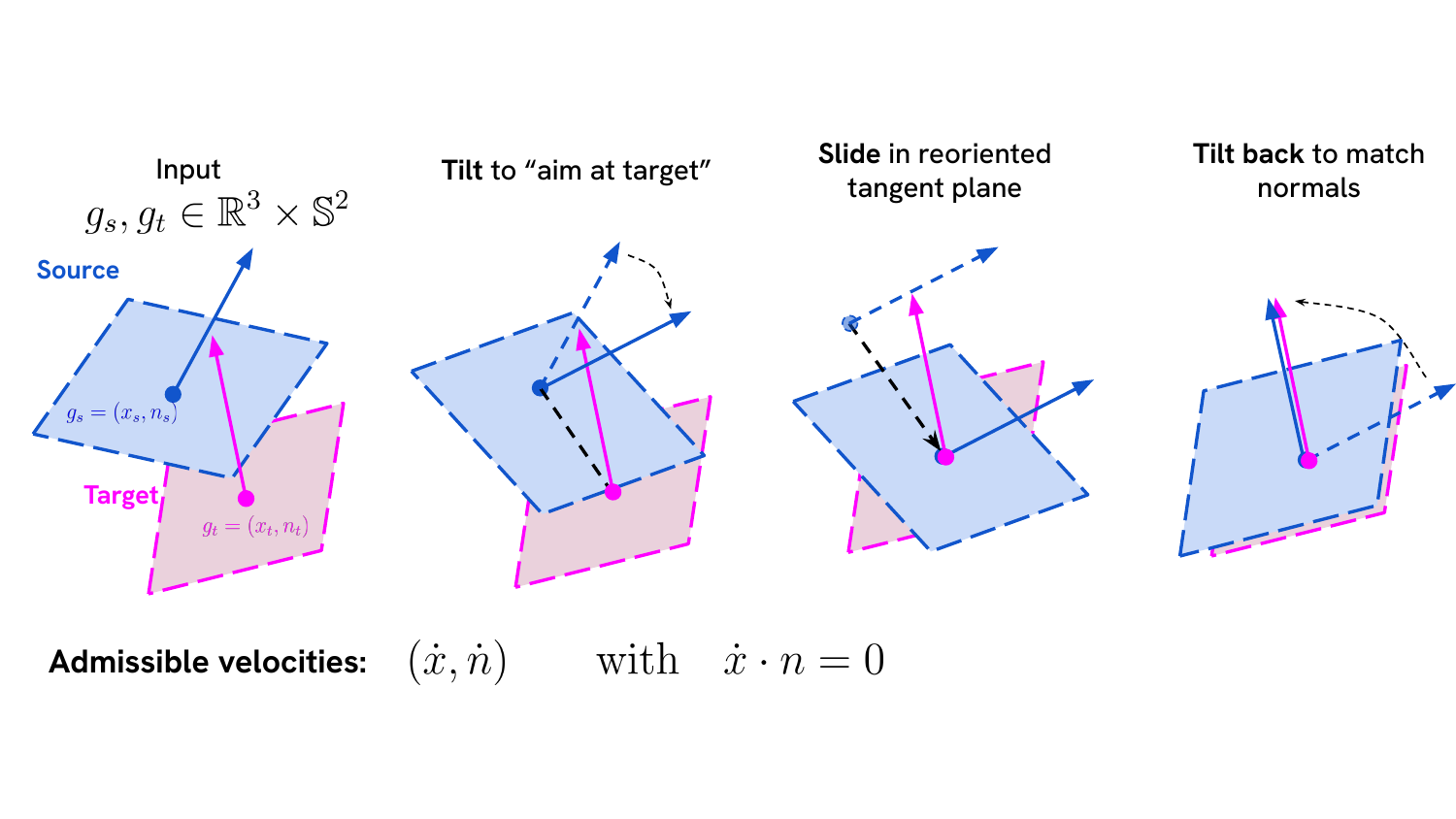}
    \caption{\textbf{Illustration of the mechanism behind the metric for matching two oriented points.} A source oriented point $g_s$ has to match a target one $g_t$. To do so, $g_s$ has to tilt first to be able to align the point position and then reorient back to match orientation. The motion has to make a detour to reach the target configuration which is exactly what a sub-Riemannian structure induces.}
    \label{fig:toy_figure_sr_idea}
\end{figure}

One way or another, this encodes the assumption that points close in $\RR^3$ are more likely to exchange information. This assumption is challenged as soon as distinct pieces of surface pass near one another: two stacked leaves in a plant, a wall, or a sheet of metal are separated by a relatively small distance and angle (opposite or co-oriented) normals. A point near an acute angle has a $k$NN ball encompassing both incident faces, folded and layered geometry stacks surfaces within any fixed radius. In all these cases Euclidean neighborhoods mix samples that should belong to distinct geometric semantic (such as for two parallel sheets). This can make denoisers merge distinct folds, segmentation labels may leak through thin structures. The standard workaround such as visibility heuristics, clustering the $k$NN by normal, hand-tuned anisotropic neighborhoods, and others treat symptoms case by case depending on the task. To address this, some approaches starts from the Euclidean neighborhood and re-weight it by normal similarity \citep{fleishman2003bilateral, guerrero2018pcpnet}, or add a normal term to the distance \citep{Rusinkiewicz2001ptp} but these are limited for describing strong orientation anisotropy. In this paper, we propose a single geometric principle to replace the Euclidean default. We lift each sample point $x \in \mathbb{R}^3$ to position-orientation space $\M=\RR^3\times\Sphere$, $g=(x,n)$ with $n$ the (estimated) unit normal, and we define the sub-bundle $\Delta\subset T\M$ as,
\begin{equation}
\label{eq:dist-intro}
\Delta_{(x,n)} \;=\; \bigl\{(\dot x,\dot n)\;:\;\inner{\dot x}{n}=0\bigr\},
\end{equation}
A metric is then built as the cost of "motion from a source to a target": we constrain these motions in $\M$ to belong in $\Delta$. This means that spatial motion must stay perpendicular to the normal of the considered point, while the latter may reorient at a cost set by a weight $w>0$. The forbidden direction is translation along the normal which is intuitively the direction in which Euclidean neighborhoods short-circuit and leak.
By tilting (reorienting the normal), sliding, and tilting back, normal displacement can still be achieved, but at an additional cost intrinsic to the metric. This is drawn from the following intuitive hypothesis: \textbf{a surface element can slide and tilt but not hop off itself} as illustrated in~\Cref{fig:toy_figure_sr_idea}.

Sub-Riemannian (SR) geometry offers a rigorous framework for this construction including in particular a strongly anisotropic metric to encode forbidden short-circuits mentioned earlier. We strongly advise the reader to read through Appendix A for a simplified background on sub-riemannian geometry and contact metric. This type of structure on roto-translation groups has demonstrated its interest for contour perception and crossing-preserving image enhancement \citep{petitot2003neurogeometry,citti2006cortical,franken2009crossing,duits2016subriemannian,boscain2014hypoelliptic}, and have recently powered equivariant PDE networks on images \citep{smets2023pde,bellaard2023analysis}. Those constructions lift curves in images, aligning spatial motion along the orientation. Here, we propose to adapt this idea to surfaces described as unstructured (oriented) point sets rather than grids.
\vspace{-2mm}
\paragraph{Contributions.} The metric structure itself is not new but its use and study as a locality prior within discrete 3D sets. In particular, this paper gather the following
\begin{enumerate}[leftmargin=1.4em,itemsep=1pt,topsep=2pt]
\item \textbf{A framework} defining a sub-Riemannian contact metric for sampled surfaces seen as position-orientation sets.
\item \textbf{A computable contact score} leveraging the Heisenberg tangent model with a key property: its square is bounded below by a quantity linear in the normal offset, uniformly in the tangential displacement. At fixed tangential displacement and small offset, the penalty is quadratic.
\item \textbf{Downstream applications} demonstrating the interest of the proposed geometric prior for learning on oriented point clouds and for heat diffusion on oriented geometric measures
\end{enumerate}

\section{Related work}
\label{sec:related}

\paragraph{Euclidean locality and its patches.} Point networks build $k$NN/radius graphs in $\RR^3$ \citep{qi2017pointnetpp,wang2019dynamic,monti2017geometric, thomas2019kpconv, wu2024ptv3}. Failures at thin structures are well known in geometry processing. To address it, bilateral weighting by normal similarity \citep{fleishman2003bilateral}, learned local shape descriptors \citep{guerrero2018pcpnet} or normal surface fitting \cite{Guennebaud2007algebraic, Öztireli2009feat} can be used to mitigate them. These approaches keep the Euclidean candidate set and re-weight it. Here, we change the locality formalism that defines the candidate set, the ranking, and the weights.

\vspace{-4mm}
\paragraph{Position-orientation and group-equivariant networks.} Group convolutions \citep{cohen2016group,bekkers2018roto,bekkers2020bspline,weiler20183d,thomas2018tensor,fuchs2020se3} lift features to (quotients of) $\SE$. PONITA \citep{bekkers2024ponita} operates on $\RR^3\times\Sphere$ and conditions messages on roto-translation invariants of point pairs: orientation is a feature. \cite{Bellaard2025RotoTranslationIM} characterize these pair invariants exhaustively and we borrow some of these. Locality, however, remains a Euclidean graph, nothing in the message function is constrained to suppress flow between incompatible orientations at first order. The same goes with \cite{Petrov2026blowup} which lifts points to the product of ambient space with the Grassmannian of tangent planes, iterates the construction so that samples differing in curvature or higher-order contact also separate. However, their lift is unoriented while ours is co-oriented (with $\mathbb{M} \cong S(T^*\mathbb{R}^3)$ and their lifted metric is a product metric with the same linear separation. 
\vspace{-4mm}
\paragraph{Sub-Riemannian image analysis.} The roto-translation lift of 2D/3D images is used in \citep{petitot2003neurogeometry,citti2006cortical}, and can further be developed for crossing-preserving diffusion \citep{franken2009crossing}, hypoelliptic inpainting \citep{boscain2014hypoelliptic}, data-driven SR geodesics \citep{bekkers2015pde,duits2016subriemannian,duits2018reedsshepp,mirebeau2018fast}, and PDE-based equivariant networks whose sharp anisotropy requires the sub-Riemannian kernel \citep{smets2023pde,bellaard2023analysis}. There, admissible spatial motion is along the lifted orientation (curves), here it is perpendicular to it. In particular, the distribution has rank 4 instead of 3. \citep{Portegies2015NewAO} give a related kernel approximation on SE(3). Our use of the nilpotentization to compute a score follows from \citep{bekkers2018nilpotent}, who group vessel key points using homogeneous norms on a nilpotent approximation of SE(3). We differ in the distribution, here defined as "motions" orthogonal to the orientation rather than along it, giving a contact structure with tangent group $\mathbb{H}^5$. The pair coordinates differs also, its is frame-free invariants rather than group-logarithm coordinates, the score is the exact distance of the tangent group rather than a homogeneous norm, and the application is also different.



\section{A sub-Riemannian structure on oriented point clouds}
\label{sec:lift}

To formalize the illustration from~\Cref{fig:toy_figure_sr_idea}, we define the state space, admissible motions and properties of contact geometry \cite{geiges2008introduction} in our context of sampled surfaces.\\
\textbf{State space. }An oriented sample is $g=(x,n)\in\M:=\RR^3\times\Sphere$. The group $\SE$ acts by $(R,t)\cdot(x,n)=(Rx+t,\,Rn)$. The action is transitive with stabilizer $\SOtwo$ (rotations about the normal), so $\M\cong\SE/\SOtwo$ is a 5-dimensional homogeneous space. A $C^2$ oriented surface $\Sigma\subset\RR^3$ lifts to the 2-dimensional sub-manifold $\tilde\Sigma=\{(x,n(x)):x\in\Sigma\}\subset\M$.\\
\textbf{Admissible motions.} An element of the state space can slide in its tangent plane and tilt its normal vector, but cannot slide along it. Reorienting by $\alpha$ costs the same as sliding $w\alpha$ along the surface.

\begin{definition}[Horizontal distribution and SR metric]
\label{def:dist}
For $w>0$ define
\begin{equation}
\label{eq:dist}
\Delta_{(x,n)}=\bigl\{(p,u)\in\RR^3\times T_n\Sphere:\inner{p}{n}=0\bigr\},
\quad
G_{(x,n)}((p,u), (q,v))= \langle p, q \rangle + w^2\langle u, v \rangle
\end{equation}
A path $t \mapsto \gamma(t)=g(t) = (x(t),n(t))$ is \emph{horizontal} if $\dot\gamma(t)\in\Delta_{\gamma(t)}$ a.e., its length is $l(\gamma)=\int\sqrt{G(\dot\gamma)}\,dt$, and the Carnot-Carathéodory distance $\dcc(g(0),g(1))$ is the minimal length of horizontal paths joining $g(0)$ to $g(1)$.
\end{definition}

$\Delta$ has rank 4 in a 5-manifold: two tangential translations and two tilts. The missing direction is translation along $n$.

\begin{proposition}[Contact structure]
\label{prop:contact}
Let $\theta_{(x,n)}(\dot x,\dot n)=\inner{\dot x}{n}$, so $\Delta=\ker\theta$. Then, $\theta$ is a contact form: $\theta\wedge(d\theta)^2$ is nowhere zero. Consequently $(\M,\Delta)$ is the unit cotangent bundle $S(T^*\R^3)$ with its canonical contact structure.
\end{proposition}

\begin{corollary}[Lifted surfaces are Legendrian]
\label{coro:legendrian}
$\tilde\Sigma$ is an integral submanifold of $\Delta$ of the maximal dimension $2$. 
\end{corollary}

\begin{proposition}[Step-2 bracket-generating]
\label{prop:bracket}
$\Delta$ is bracket-generating of step 2: $\Delta+[\Delta,\Delta]=T\M$. Consequently, because of Chow-Rashevskii theorem, any two points of $\M$ can be joined by a horizontal path and $\dcc$ is a finite metric inducing the manifold topology.
\end{proposition}
This means that the (hard) constraint cannot be integrated but the forbidden direction can be reached, at a higher cost that we will now quantify and explain in more details.

\section{A computable local contact score}
\label{sec:theory}


In the following, $P, Q$ are oriented points with $x_P, x_Q$ their respective position in $\mathbb{R}^3$ and $n_P,n_Q$ their attached normal vectors.  The true Carnot-Carathéodory distance $d_{CC}$ is
\begin{equation}
    d_{CC}(P,Q) = \inf_{\gamma}\int_0^1 \sqrt{\| \dot{x}\|^2 + w^2 \| \dot{n}\|^2} \text{d}t, \qquad \inner{\dot{x}}{n}=0
    \label{eq:cc}
\end{equation}
This is a true metric but obtaining it require solving an optimal control problem: it has no global closed-form solution but it can be locally approximated.
In this section, we (i) define a pair $(P,Q)$ through rigid invariants, (ii) replace $d_{CC}$ by the exact distance of its tangent model.

\subsection{Pair invariants}
Let's first consider the four rigid-motion invariants 
\begin{equation}
 \|r\|=\|x_Q-x_P\|,\quad \alpha=\angle(n_P,n_Q)\in[0,\pi],\quad
 \zeta_P=r\!\cdot n_P,\quad \zeta_Q=r\!\cdot n_Q,
 \label{eq:raw-invariants}
\end{equation}
and set
\begin{equation}
 \zeta=\tfrac12(\zeta_P+\zeta_Q)=\left\langle r, \frac{n_P + n_Q}{2} \right\rangle\qquad
 \rho^2=\|r\|^2-\zeta^2
 \label{eq:adapted-invariants}
\end{equation}
Then, $\zeta$ is the symmetric weight-two contact coordinate (the normal offset) and $\rho$ is the tangential spatial separation. Note that $\zeta=0$ means the vector $r$ makes equal and opposite angles with the tangent planes.

\subsection{The Heisenberg tangent model}\label{subsec:heis-tangent-model}
Mitchell's theorem \citep{Mitchell1985OnCM} identifies the metric tangent cone of a sub-Riemannian manifold at g with the Carnot group given by the nilpotentization of $(\Delta, G)$ at g. Since $\Delta$ is a contact distribution, this group is a Heisenberg group which in our case is five-dimensional $\mathbb{H}^5$. In this space, horizontal displacements of norm $a$ and vertical coordinate $\tau$ projects to a circular arc of radius $R$ and angle $\phi \in (0, 2 \pi)$, $\tau$ being the area between the arc and its chord: $a = 2R\sin{(\phi / 2)}, |\tau| = \frac{1}{2}R^2(\phi - \sin{\phi})$ and length $R \phi$ which gives the squared Heisenberg distance $H$ as
\begin{equation}
 H(a^2,\tau)= R^2 \phi^2 =a^2\left(\frac{\phi}{2\sin(\phi/2)}\right)^2 = \frac{2|\tau| \phi^2}{\phi - \sin \phi}
 \label{eq:heis-distance}
\end{equation}
where $\phi$ is the unique solution of
\begin{equation}
 \frac{\phi-\sin\phi}{1-\cos\phi}
   =\frac{4|\tau|}{a^2} = s
 \label{eq:theta}
\end{equation}
which is an isoperimetric problem. From this, we can define a "\textbf{detour cost}" $q$ such that
\begin{equation}
    q(s):= \frac{\phi(s)^2}{4\sin^2(\phi(s)/2)} = 1 + \frac{3}{4}s^2 - \frac{9}{80}s^4 + O(s^6)
    \label{eq:detour-cost}
\end{equation}

To specialize the tangent model to our metric, set $a^2 := \rho^2 + w^2\alpha^2, \tau := w\zeta$.
Both terms in $a$ have length units, while $\tau$ has length-squared units, as required by the weight-two Heisenberg scaling. 

\begin{proposition}[The Heisenberg distance approximates $d_{CC}$]
\label{prop:approx_Heisenberg}
Fix $P$ and let $Q \rightarrow P$ and $\epsilon = \|r_T\| + w\alpha + \sqrt{w|\zeta|}$ with $r_T$ the projection of $r$ onto $n^{\perp}_P$ such that horizontal coordinates are $O(\epsilon)$ and the missing normal-translation coordinate is $O(\epsilon^2)$. Up to the sign convention for the central Heisenberg coordinate, $(a, \tau)$ defined above are the horizontal norm and weight-two coordinate of the $\mathbb{H}^5$ nilpotentization to leading order. Consequently,
\begin{equation*}
    d_{CC}(P,Q)^2 = H(a^2, w\zeta) + o(\epsilon^2)
\end{equation*}
\end{proposition}
We then define the proposed \textbf{local contact score} $\ell_w$ such that
\begin{equation}
\ell_w(P,Q)^2=H(a^2,w\zeta) = a^2q\left(\frac{4|\tau|}{a^2}\right)
 \label{eq:local-score}
\end{equation}
From this, as $a \rightarrow 0$ (meaning from $P$ to $Q$, normals are co-oriented and displacement is purely normal), we have $\ell_w^2 \rightarrow 4\pi w|\zeta|$ which means a pure normal offset of "size" $\delta$ costs $\sqrt{4\pi w\delta}$. $\ell_w$ is symmetric, $\SE$-invariant and vanishes when $P \rightarrow Q$ but the triangle inequality is not verified in general which means $\ell_w$ is a semi-metric. For unoriented normals, we extend the score as
\begin{equation}
 \ell_w^{\mathrm{proj}}((x,[n]),(y,[m]))
 =\min_{s\in\{-1,1\}}\ell_w((x,n),(y,sm)),
 \label{eq:quotient}
\end{equation}

Beyond the asymptotic property of Proposition~\ref{prop:approx_Heisenberg}, we derive non-asymptotic properties of the score, ensuring it is motivated in general. 
The growth in the normal offset depends on the regime: for small offsets, the score is $l^2_w = a^2 + 12w^2\zeta^2/a^2 + \dots$ which is quadratic, and only a pure normal offset ($a=0$) gives the linear value $4\pi w |\zeta|$. For every $a$, we have $\max\{a^2, 2\pi w|\zeta|\} \leq l^2_w \leq a^2 + 4\pi w |\zeta|$, while purely quadratic penalty such as PTP adds only $O(\zeta^2)$. This lower bound separates surfaces.
We also study the, although ideal, useful case of stacked sheets with co-oriented and opposing normals. Details are given in Appendix B.


\section{Experiments}
\label{sec:experiments}
We introduced this contact score as a geometric prior to increase the separation power in "contact-prone" contexts. Supporting this claim with practical applications beyond synthetic evaluation is tricky, but we report a set of results in this section. Complementary details and explorations can be found in Appendix C. In~\Cref{subsec:synthetic}, we show how the strong anisotropy induced by the metric allows to address structural limitations when defining neighborhood in contact-prone structures. Then, in the following~\Cref{subsec:downstream_app}, we show that replacing the neighborhood mechanism in an established point-based neural network can improve its capabilities on these structures. Finally, we go further and use the kernel induced by the metric as a "similarity score" in a recent algorithm to define a diffusion process on weighted oriented discrete point sets (also called discrete varifolds \citep{charon2013varifold, kaltenmark2017varifold, buet2016varifold} in the shape analysis literature), leveraging orientation in a more principled way than using it as feature. We denote the proposed contact metric structure (\textsc{CONTACT}). For an ordered pair $i=(x_i,n_i)$, $j=(x_j,n_j)$, we write $r=x_j-x_i$, $\alpha=\angle(n_i,n_j)$, $\zeta_i=\langle r,n_i\rangle$, $\zeta_j=\langle r,n_j\rangle$, $\zeta=\tfrac12(\zeta_i+\zeta_j)$ and $\rho^2=\|r\|^2-\zeta^2$. 
\vspace{-3mm}
\paragraph{Baseline methods.} To support its interest compared to classical ways to define "locality" in oriented point sets, we compare to the following baselines: euclidean \textsc{EUC}, bilateral \citep{fleishman2003bilateral} \textsc{BIL} (which is Euclidean neighbors whose features are weighted by $e^{-\alpha^2/2\sigma_{\alpha}^2}$ before max-pooling), product metric \textsc{PROD} and point-to-plane \citep{Rusinkiewicz2001ptp} \textsc{PTP}.
\vspace{-3mm}
\paragraph{"Ablation" of the score.} To test whether the exact Heisenberg formulation is needed for the learning gains, we also replace the grouping score by two functions of the same invariants formalized in~\Cref{tab:scores}.
LIN-$\lambda$ preserves linear growth in the normal offset controlled by a parameter $\lambda$, whereas QUAD uses a quadratic penalty. With $\lambda=4\pi$, it is the exact upper envelope of \textsc{CONTACT}. 

\begin{table}[t]
\caption{\textbf{Chosen baselines to define different notions of locality in oriented point sets.} The "Off-plane behavior" column states how a score depends on displacement along the normal, which is the direction Euclidean neighborhoods short-circuit. 
All scores are $SE(3)$-invariant, symmetric and definite. 
\textsc{Bil} is applied as a weight on a Euclidean computation and the weight depends on the angle $\alpha$ through the scale $\sigma_{\alpha}=0.4$.}
\label{tab:scores}
\centering
\small
\setlength{\tabcolsep}{5pt}
\renewcommand{\arraystretch}{1.35}
\begin{tabular}{llc}
\toprule
Score & Squared pairwise score & Off-plane behavior \\
\midrule
\textsc{Euc}     & $\|r\|^2$                                          & ---                      \\
\textsc{Bil}     & $\|r\|^2$, weight $\times\,e^{-\alpha^2/2\sigma_\alpha^2}$ & ---              \\
\textsc{Prod (section)}    & $\|r\|^2+w^2\|n_i-n_j\|^2$                         & product penalty                     \\
\textsc{Prod (angular)}    & $\|r\|^2 + w^2\alpha^2$                         & product penalty                     \\
\textsc{Ptp}     & $\|r\|^2+ 2\pi\big(\zeta_i^2+\zeta_j^2\big)$        & quadratic off-plane    \\
\midrule
\textsc{LIN-$\lambda$} & $\rho^2+w^2\alpha^2+\lambda w|\zeta|$ & symmetrized, linear \\
\textsc{QUAD} & $\rho^2+w^2\alpha^2+4\pi\zeta^2$ & symmetrized, quadratic \\
\textsc{Contact} & $H\big(\rho^2+w^2\alpha^2,\,w\zeta\big)$ & symmetrized, quadratic to linear \\
\bottomrule
\end{tabular}
\end{table}

\subsection{Synthetic validations}
\label{subsec:synthetic}

\begin{figure}[ht!]
  \centering
  \includegraphics[width=0.95\linewidth]{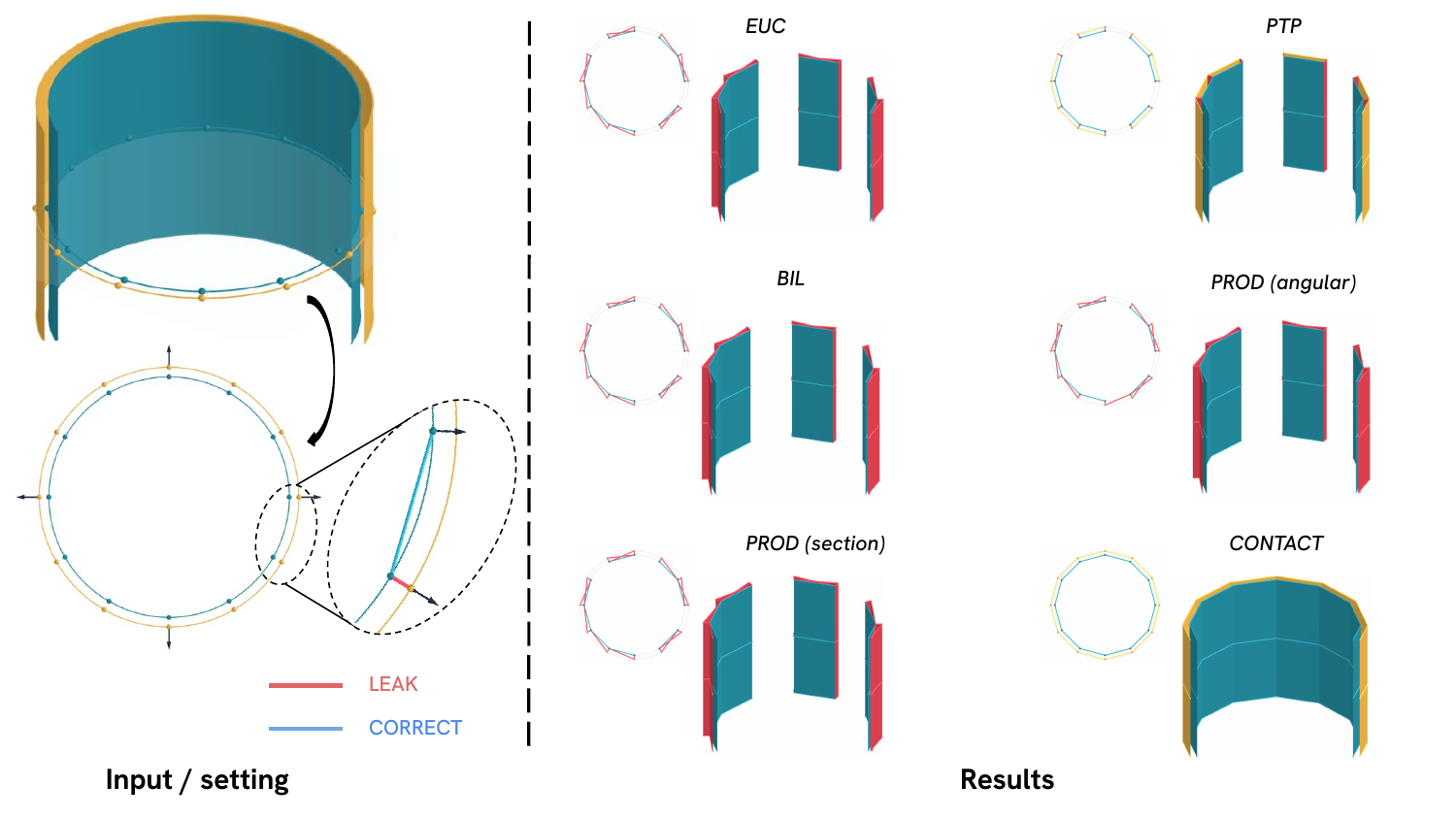}
  \vspace{-4mm}
  \caption{\textbf{Setting and results for the cylinder reconstruction experiment.} For each method, we take the undirected union of its two-neighbor graph and extrude every edge into a triangulated strip. Red marks cross-cylinder edges (leaks) and false strips. CONTACT is the only among compared approach to exhibit two separate polygonal cylinders.}
  \label{fig:synthetic-reconstruction}
\end{figure}

We start by exhibiting the type of setting in which the contact structure we propose is the most relevant. We consider two coaxial cylinders of radii $R$ and $R+\delta = 1.08R$ (very close) and sample 12 equally spaced points from each with co-oriented normals. The task is, when selecting a point, computing its 2 closest neighbors (on each side). Such a neighbor is correct if it belongs to the cylinder of the queried point. We compare the resulting reconstruction and showcase in~\Cref{fig:synthetic-reconstruction} how the CONTACT score is the only approach among the mentioned baselines to correctly reconstruct the two cylinders. Regarding the linear and quadratic approximation of \textsc{CONTACT}, \textsc{LIN}-$\lambda$ fails for $\lambda=2\pi$ but succeeds for $\lambda=4\pi$.
In particular, we see that the cylinders are separated by the weight-two coordinate $\zeta \approx \delta$. Note that if the cylinders would have been with opposing normals, $\zeta=0$ and $\ell_w^2= \rho^2 + w^2\alpha^2 \geq \pi^2w^2$, independently of thickness. This will be even more visible with the heat diffusion experiment in~\ref{sec:oriented-diffusion}. We also mention that this synthetic experiment is pushed further in Appendix C with effect of normal noise and staggered sampling.

\subsection{Downstream learning applications}
\label{subsec:downstream_app}

We consider two tasks: segmentation and shape correspondence. For both, we employ a PointNet++ \citep{qi2017pointnetpp} model, widely used for point cloud processing. It is notably relying on euclidean neighborhood aggregation to produce a hierarchical pooling of features to better encode structure. We replace this aggregation with the baselines and report the final metrics.

\textbf{Plant segmentation.}
\label{sec:pheno4d}
\textit{Data.}
Pheno4D \citep{schunck2021pheno4d} is a set of high-resolution laser scans of growing maize and tomato plants. We use every annotated scan: $126$ point clouds over $14$ plants ($7$ maize $\times\,7$ days, $7$ tomato $\times\,11$ days), each carrying $1.5$--$4.5$ million points. 
Both use the convention $0=$ soil, $1=$ stem, $2,3,\ldots=$ leaf instances. We drop the soil, which is $85\%$ of every scan and separable by a height threshold, and keep the plant. Each cloud is subsampled to $2^{14} = 16{,}384$ points, equipped with unoriented normals estimated from local PCA over $24$ neighbors, and normalized to the unit sphere. Training draws $4{,}096$ points per cloud. Because the normals are unoriented, we use~\eqref{eq:quotient}. 

\textit{Task.}
Binary segmentation on the plant: \textbf{stem or leaf}. Train/test splits are by plant, since the scans of one plant are the same plant on consecutive days and are not independent. Maize06, Maize07, Tomato06 and Tomato07 are held out for test, Maize05 and Tomato05 for validation, and the remaining 8 plants are used for training.

\textit{Method.}
We train a PointNet++ segmentation network
\citep{qi2017pointnetpp} with four set-abstraction levels at $1024$, $256$, $64$ and $32$ centroids, $k=32$ neighbors per centroid, a feature-propagation decoder, and a two-class head (leaf or stem). The network trained for 100 epochs, replacing only the grouping mechanism. Normals are an input feature in each variant, including \textsc{Euc}, so the baseline sees exactly the same information. All variants use $k$-nearest-neighbor grouping rather than a ball query, because a radius is not comparable across scores with different units. Three seeds are run. More training details in Appendix C.

\textit{Results.}
In Table~\ref{tab:pheno4d}, we report the test metrics. We highlight how using the proposed contact neighborhood increase the model's ability to segment in this context. 


\begin{table}[ht!]
\centering\small
\begin{tabular}{lcccc}
\toprule
grouping & mIoU & accuracy & med. $|\zeta|/\|r\|$ & normal dev. \\
\midrule
\textsc{Euc} & $0.750\pm0.027$ & $0.913\pm0.014$ & 0.0680 & 12.62°\\
\textsc{Bil} & $0.789 \pm 0.026$ & $0.929 \pm 0.014$ & 0.0665 & 11.93°\\
\textsc{Prod} (Section) & $0.788\pm 0.030$ & $0.930 \pm 0.013$ & 0.0666 & 11.96° \\
\textsc{Ptp} & $0.790\pm0.011$ & $0.933\pm0.006$ & 0.0574 & 11.44°\\
\midrule
LIN-$2\pi$ & $0.827 \pm 0.018$ & $0.944 \pm 0.009$ & 0.0476 & 10.38° \\
LIN-$4\pi$ & $0.797 \pm 0.023$ & $0.936 \pm 0.009$ & - & - \\
QUAD & $0.816\pm0.008$ & $0.942\pm0.003$ & 0.0551 & 10.08°\\
\textsc{Contact} & $\mathbf{0.835 \pm 0.023}$ & $\mathbf{0.950 \pm 0.007}$ & 0.0484 & 10.14°\\
\bottomrule
\end{tabular}
\caption{\textbf{Pheno4D stem/leaf segmentation.} We report mean $\pm$ standard
deviation over three seeds. Only the grouping geometry differs between rows. In the last two columns, we report how much normals agree on computed patch (from the same queried points) on average over ten point clouds.}
\label{tab:pheno4d}
\end{table}


\textit{Analysis.}
What the grouping changes is the shape of the patch. We measure the weight-two coordinate itself over the neighbors each score selects through $|\zeta|/\|r\|$, which express how far off the query's own tangent plane each grouped point lies. We do this over ten clouds and report the average in the last two columns of~\Cref{tab:pheno4d}.
We observe that the columns are almost monotone together. The contact score selects patches much "flatter" than Euclidean grouping and point-to-plane, the local MLP is handed a cleaner tangential neighborhood. Of course, correlation does not imply causation but we still report the trend. 

\begin{figure}[t]
\centering
\includegraphics[trim={0 2.4cm 0 2.4cm},clip,width=0.95\linewidth]{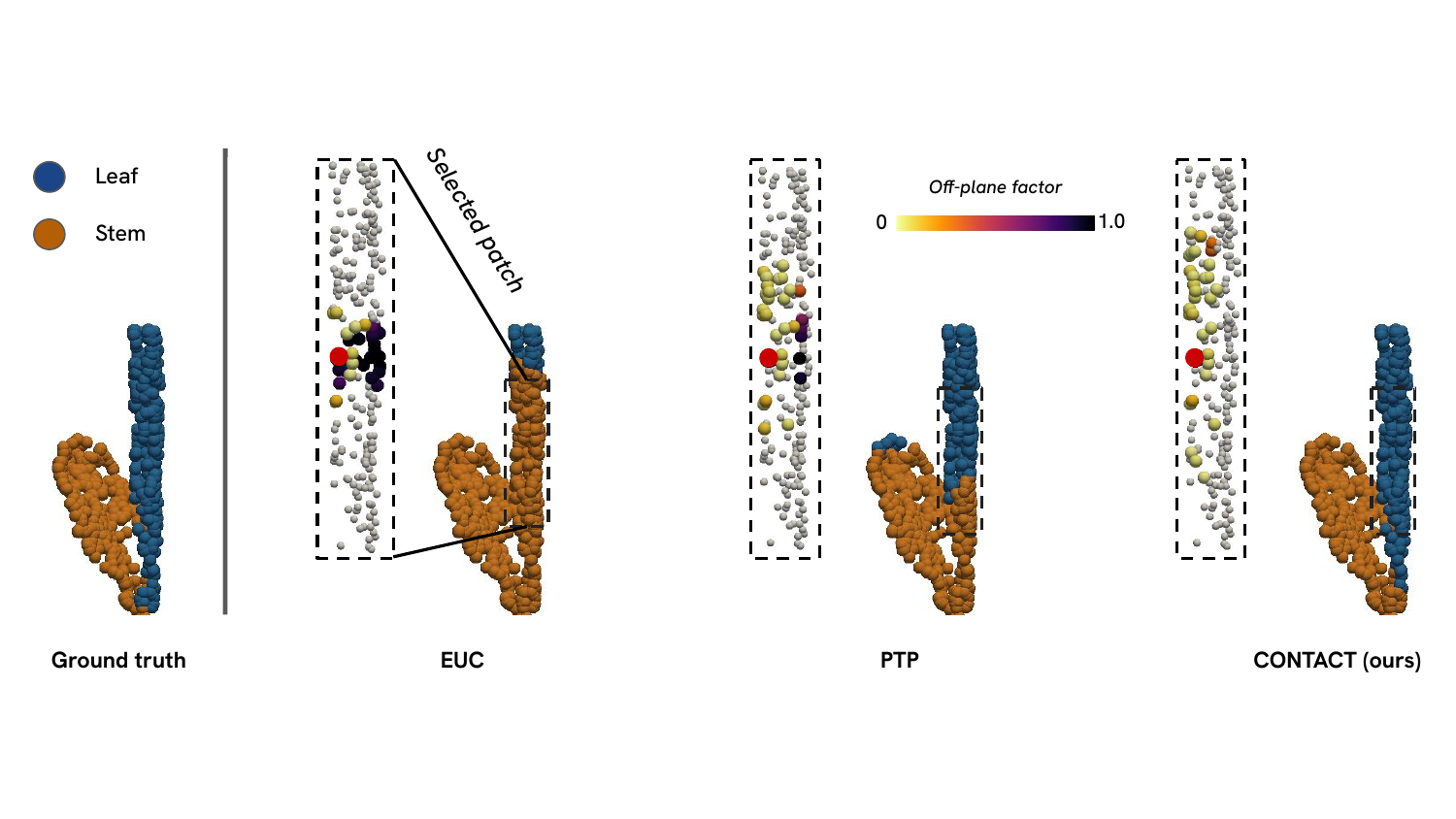}
\vspace{-3mm}
\caption{\textbf{Qualitative focus on the Pheno4D segmentation task.} For each method, we show a selected patch from the chosen neighborhood formulation and color points depending on the normalized symmetric offset $|\zeta|/\|r\|$.}
\label{fig:pheno4dnbr}
\end{figure}


\textbf{Dense shape correspondence on FAUST.} 
\textit{Data.}
FAUST \citep{Bogo2014faust} provides $100$ registered human body meshes: $10$ subjects in $10$ poses each, each one with the same $6890$-vertex topology. Vertex $i$ is therefore the same semantic point on every shape, which gives dense ground-truth correspondence with no annotation step and no ambiguity. For this one, we either estimate normals by local PCA as before or we take the true oriented normals from the meshes.

\textit{Task and metric.}
Given a human mesh, label every point with the template vertex it corresponds to: a $6890$-way per-point classification. The metric is the geodesic distance along the template surface between the predicted and the true vertex, normalized by the geodesic diameter. We report its mean geodesic error overall and focused on a "contact stratum": points at which some vertex within three sampling scales is more than 0.3 geodesic unit away (normalized scale) along the surface. This concerns 385 test points per mesh, and it is where the grouping geometry matter more. Geodesic distances are precomputed once on the template with the heat method \citep{crane2013geodesics}.




\textit{Method.}
The network and protocol are similar to the plant segmentation experiment: a single-scale-grouping PointNet++ \citep{qi2017pointnetpp} with four set-abstraction levels ($1024$, $256$, $64$,$32$ centroids, $k=32$), a feature-propagation decoder, and a $6890$ head.
Splits are by identity, test $8$-$9$, validation $7$, training $0$-$6$. Three seeds are run.

\textit{Results.} We report metrics in~\Cref{tab:faust_combined} and we observe that overall geodesic error is improved with the CONTACT neighborhood computation with PCA normals. With oriented normals, the differences are less significant and within one standard deviation.

\begin{table}[ht!]
\centering\small
\begin{tabular}{llcc}
\toprule
normals & grouping & geodesic error & \textbf{contact stratum}\\
\midrule
\multirow{7}{*}{unoriented (PCA)}
 & \textsc{Euc}     & $0.0137\pm0.0010$          & $0.0828\pm0.0110$  \\
 & \textsc{Bil}     & $0.0129\pm0.0015$          & $0.0675\pm0.0122$ \\
 & \textsc{Prod (Section)}  & $0.0132\pm0.0009$    & $0.0701\pm0.0080$ \\
 & \textsc{Ptp}    & $0.0131\pm0.0008$  & $0.0740\pm0.0126$ \\ 
 & LIN-$2\pi$ & $0.0126 \pm 0.0002$ & $0.0668 \pm 0.0090$ \\
 & LIN-$4\pi$ & $0.0125 \pm 0.0002$ & $0.0625 \pm 0.0091$ \\
 & QUAD & $0.0128 \pm 0.0006$ & $0.0670 \pm 0.0076$ \\
 & \textsc{Contact} & $\mathbf{0.0121\pm0.0007}$ & $\mathbf{0.0568\pm0.0073}$ \\
\midrule
\multirow{7}{*}{oriented (GT)}
 & \textsc{Euc}     & $0.0098\pm0.0005$          & $0.0434\pm0.0085$  \\
 & \textsc{Bil}     & $0.0102\pm0.0004$          & $0.0476\pm0.0072$ \\
 & \textsc{Prod (Section)}  & $0.0097\pm0.0004$ & $0.0414\pm0.0086$   \\
 & \textsc{Ptp}     & $0.0100\pm0.0006$          & $0.0438\pm0.0056$ \\
 & LIN-$2\pi$ & $0.0097 \pm 0.0005$ & $0.0390 \pm 0.0040$ \\
 & LIN-$4\pi$ & $0.0098 \pm 0.0008$ & $\mathbf{0.0383 \pm 0.0096}$ \\
 & QUAD & $0.0098 \pm 0.0003$ & $0.0392 \pm 0.0063$ \\
 & \textsc{Contact} & $\mathbf{0.0095\pm0.0004}$ & $0.0385\pm0.0050$ \\
\bottomrule
\end{tabular}
\caption{\textbf{FAUST dense correspondence on test subjects.} We report mean $\pm$ standard deviation over three seeds. Geodesic error is normalized by the geodesic diameter.}
\label{tab:faust_combined}
\end{table}


In particular, we see that the effect is localized to the contact stratum. With PCA normals, against Euclidean grouping, the contact-stratum error drops from 0.0828 to 0.0568. 

Interestingly, for both experiments (segmentation and correspondence), we see a trend supporting the use of the invariants from~\eqref{eq:adapted-invariants} as a strong geometric prior for learning. In particular, the strong performance of \textsc{LIN} supports that the linearity in $|\zeta|$ matters.

\subsection{Heat diffusion on oriented point clouds}
\label{sec:oriented-diffusion}

Beyond neighborhoods, the score $\ell_w$ defines a similarity kernel we can use to simulate diffusions with the algorithm proposed in \cite{kessler2026sinkhorn}: it builds diffusion operator from \textbf{similarity kernels}. We represent an oriented point cloud by the discrete measure $\mu_N=\sum_i m_i\delta_{P_i}$, where $P_i=(x_i,n_i)$ and $m_i>0$ is an area weight. A temperature signal $u_i$ carries total heat $\sum_i m_i u_i$. We then define a \textbf{contact similarity kernel} as $K_{ij}=\exp[-\ell_w(P_i,P_j)^2/(2\sigma^2)]$ using the contact score.
\vspace{-3mm}
\paragraph{From similarities to diffusion.}
The symmetric Sinkhorn algorithm of \citet{kessler2026sinkhorn} repeatedly adjust one positive scale factor per sample until each row of the averaging operator sums to one. Applying the same factors at both ends of each interaction preserves symmetry with respect to the area weights. With $M=\operatorname{diag}(m_i)$, the updates and resulting operator are
\begin{equation}
  \lambda_i\leftarrow
  \sqrt{\frac{\lambda_i}{\sum_j K_{ij}m_j\lambda_j}},
  \qquad Q=\Lambda K\Lambda M,
  \qquad \Lambda=\operatorname{diag}(\lambda_i).
  \label{eq:oriented-sinkhorn}
\end{equation}
At convergence, $Q\mathbf{1}=\mathbf{1}$ and $MQ=Q^TM$: averaging preserves both a uniform temperature and the integrated heat. The CONTACT kernel does not need to be positive semi-definite, so its balanced operator can have negative eigenvalues. We therefore construct continuous heat flow as
\begin{equation}
  A=\frac{Q-\operatorname{diag}(Q\mathbf{1})}{s},
  \qquad u(t)=\exp(tA)u(0),
  \qquad s=\frac{\sigma^2}{2}.
  \label{eq:oriented-heat}
\end{equation}
Non-negative off-diagonal rates and symmetry ensure positive temperatures and heat conservation: at exact convergence, $A=(Q-I)/s$. This continuous-time extension is distinct from the native Gaussian baseline $u_G(k s)=Q_G^k u(0)$, obtained by replacing $\ell_w$ with Euclidean distance. Here $s$ is the nominal Gaussian time scale, without fitting diffusion speeds. The comparison below evaluates the complete diffusion constructions used for the CONTACT and Gaussian kernels.
\vspace{-3mm}
\paragraph{Validation.}
We compare with the point-cloud Laplacian of \citet{Sharp2020ALF} and ordinary cotangent mesh heat as reference, each evolved as $\exp(-tM^{-1}L)u(0)$. We evaluate on 3 shapes: the Stanford bunny and dragon and a set of 3 stacked sheet in which two are co-oriented and the last one has opposite normals. We measure the normalized discrepancy as follows: let $u_i(t)$ be the heat field computed by a given method and $u_i^{mesh}(t)$ the cotangent mesh reference, $m_i$ is still the lumped mesh vertex mass. The discrepancy $D$ is then
\begin{equation}
    D(t) = \frac{1}{2}\sum_i \left | \frac{m_iu_i(t)}{\sum_j m_ju_j(t)} - \frac{m_iu_i^{mesh}(t)}{\sum_j m_ju_j^{mesh}(t)} \right |
\end{equation}


\begin{figure}[ht!]
  \centering
  \includegraphics[width=1.0\linewidth]{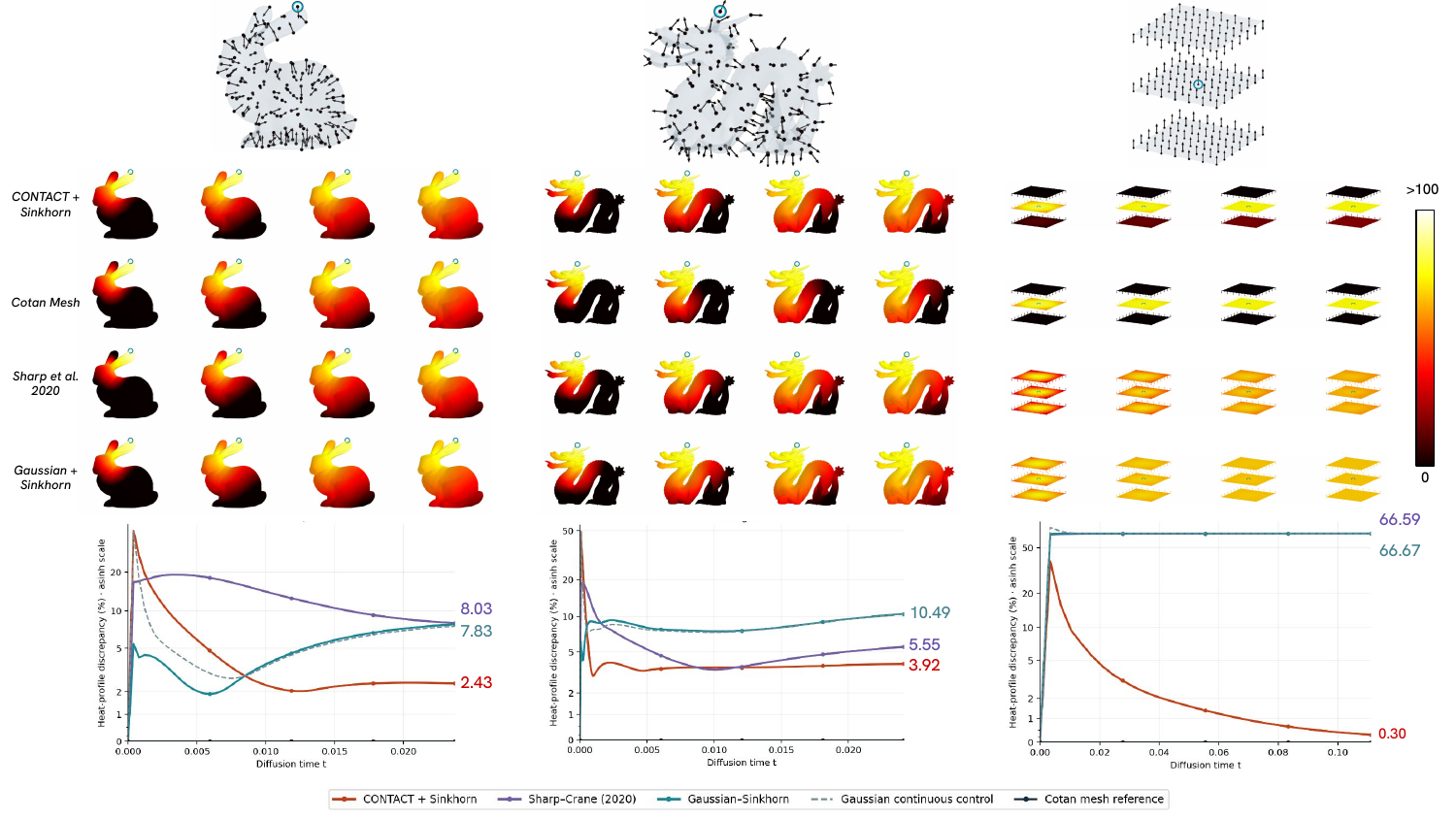}
  \caption{\textbf{Heat diffusion comparison.} We compare the computed heat diffusion process against a reference mesh diffusion (from the cotangent Laplacian). On the graphs is reported the normalized heat-profile discrepancy $D(t)$ (in \%) which is the total variation measured with common mesh areas. Reported numbers are the final discrepancy. Lower means closer to the mesh diffusion.}
  \label{fig:oriented-sheets}
\end{figure}

Results are reported in~\Cref{fig:oriented-sheets}. In these three examples, the CONTACT construction substantially reduces leakage and diffusion shortcuts across nearby disconnected surfaces and achieves the best final normalized heat-profile agreement, while exhibiting a pronounced short-time mismatch with cotangent mesh diffusion. An additional example and a wave propagation simulation with a similar approach are given in the appendix.

Under a common rule (fixed steps $Q^k$ for both kernels), the proposed \textsc{CONTACT} score stays more accurate at the final time on all three shapes and at all times on the stacked sheets. The early spikes observed on the curved shapes comes from the continuous rule, which keeps at least $e^{-t/s}$ of the heat at the source. With a smooth start, this early discrepancy drops significantly (around 5\%). The linear approximation of the score \textsc{LIN}-$\lambda$ is able to separate the stacked sheets but is much worse on curved shapes.

\section{Discussion, limitations, and conclusion}
\label{sec:discussion}

\paragraph{Limitations.} While we demonstrated several advantage of using a sub-Riemannian contact geometry to describe locality in oriented point clouds, we still mention limitations. First, normals are an input and the approach is sensible to noise as additional results with synthetic data reported in Appendix C and its effect should be studied in depth in future work. Similarly, the weight $w$ is fixed but a locally adapted or learned computation is a natural extension. Also if the sampling is not dense enough, we drift from the "local approximation" regime where the Heisenberg distance can be used to approximate the true Carnot-Carathéodory distance.

\paragraph{Conclusion and future work.} We introduced a metric structure on oriented point clouds to reduce shortcuts and leakage across contact when connecting neighbors. We showcased its use as a drop-in replacement of local aggregation operations in both deep learning framework and PDE computation (heat diffusion). This resulted in better handling of contact-prone structures without relying on a clean connectivity matrix. From this, many directions are worth pursuing, either on the theoretical side with potential extensions of the metric to textured shapes, curvature driven contact similar to \cite{Petrov2026blowup} but with stronger anisotropy. Or, it could be used in applications involving normals such as refining mesh reconstruction algorithms from Gaussian splatting using normal maps \citep{gomez2026blobsspokeshighfidelitysurface} or defining hypoelliptic heat-kernel signatures as pose-invariant descriptors generalizing \citet{sun2009concise} as a purely geometric feature descriptor on oriented point clouds. Either way, we hope this work will open ways to improve geometry processing and learning algorithms on oriented sets in general. 

\clearpage

\section*{AI use statement}
We used generative AI tools for two things: improving language and readability of the manuscript and deign and run of synthetic experiments in order to control parameter tuning. We reviewed all AI-assisted edits and take full responsibility for the final content of this work.

\section*{Reproducibility statement}
The main paper and appendix provide the problem formulation, model architecture, training objective, inference strategy, evaluation protocols, metric definitions, and ablations. Code will be made available after the review process. The datasets used (Pheno4D \cite{schunck2021pheno4d}, FAUST \cite{Bogo2014faust}) are publicly available, with the splits given in Section 5.2. The synthetic shapes are generated by from script in the code and will be made available along with it.

%% file: supplementary.tex
\appendix

\section{Background on sub-Riemannian geometry and contact structure}
\label{app:sr-background}

On a high level, \textbf{sub-Riemannian} geometry is a metric space, in which, at every point, you are allowed to move only in certain directions: there is a hard constraint on velocity directions. 
\subsection{Definition}
Let $M$ be a smooth manifold. A distribution $\Delta$ assigns to each point $p$ a linear subspace $\Delta_p\subseteq T_pM$: the directions we are allowed to move in at $p$. A curve $\gamma$ is \textbf{horizontal} if $\dot\gamma(t)\in\Delta_{\gamma(t)}$ for almost every $t$. A sub-Riemannian structure is a distribution with an inner product $G$ on it. Note that $G$ is defined only on $\Delta$. If $\Delta=TM$, the geometry is Riemannian. The length of a horizontal curve is $\ell(\gamma)=\int\sqrt{G(\dot\gamma,\dot\gamma)}\,dt$ as usual. The \textbf{Carnot-Carathéodory distance} is
\begin{equation}
\dcc(p,q)=\inf\{\ell(\gamma):\gamma \text{ horizontal},\ \gamma(0)=p,\ \gamma(1)=q\}.
\label{eq:cc0}
\end{equation}

Here, different from Riemannian geometry, there may be no horizontal curve joining $p$ to $q$, in which case the infimum is over an empty set and $\dcc=+\infty$. Even when curves exist, the distance does not necessarily behave like a Riemannian one at small scales. However, in our case, we do not see such degenerate case thanks to the following theorem.

\subsection{Brackets and Chow-Rashevskii}

For two smooth vector field $X, Y: M \rightarrow TM$, the Lie bracket is defined as
\begin{equation}
    [X, Y](f) = X(Y(f)) - Y(X(f)) \quad \text{for all } f\in C^{\infty}(M)
\end{equation}
The distribution $\Delta$ is itself a vector field. We can write $\Delta^1=\Delta$ and $\Delta^{k+1}=\Delta^k+[\Delta,\Delta^k]$. The distribution is \textbf{bracket-generating} if $\Delta^k_p=T_pM$ for some finite $k$ at every $p$, and the smallest such $k$ is the \textbf{step}. A step-two distribution is one where a single bracket suffices, which is the case here and the simplest interesting one.

\begin{theorem}[Chow-Rashevskii]
If $\Delta$ is bracket-generating on a connected manifold $M$, then any two points are joined by a horizontal curve, $\dcc$ is finite, and it induces the manifold topology.
\end{theorem}
Basically, if the theorem is verified, distance is defined everywhere, you just might have to make a detour to get there.

\subsection{Ball-box}
As explained just before, when it is bracket-generating, sub-riemannian structures induce a metric defined everywhere on the space but, because some directions are forbidden, connecting two points may require to take a (potentially large) detour. Concretely, assign to each direction a weight: one if it lies in $\Delta$, two if it
first appears in $\Delta^2$, and so on. In a step-two structure with $\Delta$ of rank $m$ inside a manifold of dimension $m+c$, there are $m$ weight-one directions and $c$ weight-two directions.

\begin{theorem}[Ball-box] \label{app:ballbox}
In suitable local coordinates, the sub-Riemannian ball of radius $\varepsilon$ is
comparable to a box of side $\varepsilon$ in each weight-one coordinate and side
$\varepsilon^{2}$ in each weight-two coordinate.
\end{theorem}
Because of this, a sub-Riemannian ball is \textbf{anisotropic at every scale}, and the anisotropy gets stronger as the scale shrinks, since $\varepsilon^2/\varepsilon\to0$ as $\varepsilon\to0$. A Riemannian manifold looks flatter and flatter as you zoom in but a sub-Riemannian one looks equally squashed at every scale, because zooming in is a symmetry of the structure. The volume of the ball scales like $\varepsilon^{m}\cdot\varepsilon^{2c}=\varepsilon^{m+2c}$, so the metric dimension is $m+2c$, strictly larger than the topological dimension $m+c$.

\subsection{Contact structures}
A contact structure is a particular case of sub-Riemannian geometry. When the distribution $\Delta$ is of corank one, it is the kernel of a one-form $\theta$. The Frobenius theorem states that $\ker\theta$ is integrable, the manifold is foliated by surfaces tangent to it, and a horizontal curve can never leave the leaf it starts on, exactly when $\theta\wedge d\theta=0$. A \textbf{contact structure} is the opposite extreme: on a $(2m{+}1)$-manifold, $\theta\wedge(d\theta)^m$ is nowhere zero. There are no leaves at all, not even locally, and the horizontal curves through a point reach a full neighborhood. Two consequences are used later. First, integral submanifolds of a contact structure have dimension at most $m$, and those attaining it are called \textbf{Legendrian}. The lift of a surface to the space of its own oriented tangent planes is then Legendrian, which in our case is intuitively describing how a surface is "perpendicular to its normal field" in this geometry. Second, a contact distribution admits no non-trivial abnormal minimizers, so every minimizing geodesic satisfies the normal Hamiltonian equations. 

\subsection{Nilpotent approximation}
The ball-box theorem said that a sub-Riemannian ball stays squashed no matter how far you zoom in. Then, if we zoom "anisotropically", matching each direction to its own weight, the limit exists and is a model space in which we can compute efficiently.\\
Concretely, fix $p$ and choose \textbf{privileged coordinates} at $p$: local coordinates centered at $p$ in which every coordinate carries a well-defined weight, one for the directions lying in $\Delta$ and two for those that
appear only after a bracket. Writing $(y,z) \in \mathbb{R}^m \times\mathbb{R}^c$ for the weight-one and weight-two coordinates of a step-two structure, the anisotropic zoom is the family of dilations
\begin{equation}
  \delta_\varepsilon (y, z) \;=\; (\varepsilon y,\, \varepsilon^2 z),
  \qquad \varepsilon > 0 ,
  \label{eq:dilation}
\end{equation}
This scaling is self-similar thanks to the ball-box theorem. Pulling the structure back through $\delta_\varepsilon$ and taking the limit $\varepsilon \to 0$ keeps the horizontal frame from degenerating: in privileged coordinates this limit discard the lowest order terms.
What remains is the \textbf{nilpotentisation} of $(\Delta, G)$ at $p$, a graded nilpotent Lie group $\mathcal{N}_p$ with its own left-invariant sub-Riemannian structure. Mitchell's theorem~\citep{Mitchell1985OnCM,bellaiche1996} formalizes this limit at a regular point so we have
\begin{equation}
  \bigl( \mathbb{M},\, \tfrac{1}{\varepsilon} d_{CC},\, p \bigr)
  \;\xrightarrow[\ \varepsilon \to 0\ ]{\ \mathrm{GH}\ }\;
  \bigl( \mathcal{N}_p,\, d_{\mathcal{N}_p},\, 0 \bigr),
  \label{eq:tangentcone}
\end{equation}
in the Gromov-Hausdorff (GH) sense (convergence of metric spaces). $\mathcal{N}_p$ plays for a sub-Riemannian manifold the role the tangent space plays for a Riemannian manifold. It is called the \textbf{metric tangent cone}. The difference is that the model is a group instead of a vector space, and it can be a different group at every point.
 
In particular, for a contact structure it is. Corank one with $\mathrm{d}\theta$ non-degenerate on $\Delta$ leaves no freedom in the graded bracket: the nilpotentisation at every point of a $(2m{+}1)$-dimensional contact manifold is the Heisenberg group $\mathbb{H}^{2m+1}$, and in our case $m = 2$ gives $\mathbb{H}^5$ everywhere. The Carnot-Caratheodory distance $d_{CC}$ has no global closed form in general, but the Heisenberg distance does, and by~\eqref{eq:tangentcone} it is the leading term of the former. for $q$ approaching $p$ in the scaling~\eqref{eq:dilation},
$d_{CC}(p,q)$ and the Heisenberg distance between the corresponding
privileged coordinates agree to first order.

For the rank-4 contact distribution of this paper there are 4 weight-one coordinates: two tangential translations and two reorientation coordinates. And one weight-two coordinate generated by their brackets.

\section{Proofs}
\label{app:proofs}

\subsection{From the main paper}

\textbf{Proposition 3.2}(Contact Structure)
Let $\theta_{(x,n)}(\dot x,\dot n)=\inner{\dot x}{n}$, so $\Delta=\ker\theta$. Then, $\theta$ is a contact form: $\theta\wedge(d\theta)^2$ is nowhere zero. Consequently $(\M,\Delta)$ is the unit cotangent bundle $S(T^*\R^3)$ with its canonical contact structure.
\begin{proof}\vspace{-3mm}
Both $n^\perp$ (the admissible $\dot x$) and $T_n\mathbb{S}^2$ (the admissible $\dot n$)
are canonically identified with the same $2$-plane $n^\perp$, so
$\Delta_{(x,n)}\cong n^\perp\oplus n^\perp$. Writing $\theta=\sum_i n_i\,dx_i$ gives $d\theta=\sum_i dn_i\wedge dx_i$, hence for horizontal vectors $d\theta\big((\dot x,\dot n),(\dot x',\dot n')\big)=\inner{\dot n}{\dot x'}-\inner{\dot n'}{\dot x}$, the canonical symplectic pairing on $n^\perp\oplus n^\perp$, which is non-degenerate. For a corank-one distribution $\ker \theta$ on a $5$ dimensional manifold (as $\mathbb{M}$, non-degeneracy of $d\theta$ on $\ker \theta$ is equivalent to $\theta \wedge(d\theta)^2 \neq 0$ \cite{geiges2008introduction}.
\end{proof}

\textbf{Corollary 3.3}(Lifted surfaces are Legendrian)
$\tilde\Sigma$ is an integral submanifold of $\Delta$ of the maximal dimension
$2$.
\begin{proof}\vspace{-3mm}
$T_{(x,n)}\tilde\Sigma=\{(\dot x,dn(\dot x)):\dot x\in T_x\Sigma\}$ and
$T_x\Sigma=n^\perp$, so $\inner{\dot x}{n}=0$ automatically. In a $(2m{+}1)$-dimensional
contact manifold integral submanifolds have dimension at most $m$ and here $m=2$.
\end{proof}

\textbf{Proposition 3.4}(Step-2 bracket-generating)
$\Delta$ is bracket-generating of step 2: $\Delta+[\Delta,\Delta]=T\M$. Consequently, because of Chow-Rashevskii theorem, any two points of $\M$ can be joined by a horizontal path and $\dcc$ is a finite metric inducing the manifold topology.
\begin{proof} \vspace{-3mm}
Fix the basis $P_1,P_2,P_3$ (infinitesimal translations) and $J_1,J_2,J_3$ (infinitesimal rotations) of the Lie algebra $\mathfrak{se}(3)$, with brackets
\[
[J_i,J_j]=\epsilon_{ijk}J_k,\qquad [J_i,P_j]=\epsilon_{ijk}P_k,\qquad [P_i,P_j]=0 .
\]
Let $\pi:\SE\to\M=\SE/\SOtwo$, $\pi(R,t)=(t,Re_3)$, where $\SOtwo=\exp(\mathbb{R}J_3)$ stabilizes $(0,e_3)$. Push-forwards of the left-invariant fields: at $g=(R,t)$, $d\pi(P_i)$ is translation by $Re_i$ and $d\pi(J_i)$ rotates $n=Re_3$ about $Re_i$; in particular $d\pi(J_3)=0$ (vertical), $d\pi(P_3)$ is translation along $n$, and $d\pi(J_1),d\pi(J_2)$ span $T_n\Sphere$.

Define the left-invariant distribution $H=\mathrm{span}\{P_1,P_2,J_1,J_2\}$ on $\SE$. Then $d\pi(H_g)=\{(\dot x,\dot n):\inner{\dot x}{n}=0\}=\Delta_{\pi(g)}$: the spatial part $\mathrm{span}\{Re_1,Re_2\}=n^{\perp}$ and the angular part is all of $T_n\Sphere$. Compute
\[
[J_1,P_2]=P_3,\qquad [J_2,P_1]=-P_3,\qquad [J_1,J_2]=J_3 .
\]
Hence $H+[H,H]\supseteq\mathrm{span}\{P_1,P_2,P_3,J_1,J_2,J_3\}=\mathfrak{se}(3)$: $H$ is bracket-generating of step 2 on $\SE$, and pushing forward, $\Delta+[\Delta,\Delta]=T\M$ (the only direction missing from $\Delta$, translation along $n=d\pi(P_3)$, is generated; $J_3$ is vertical). Chow-Rashevskii \citep{chow1939systeme,rashevskii1938connectability,montgomery2002tour} gives horizontal connectivity and that $\dcc$ is a finite metric for the induced topology.
\end{proof}

\textbf{Proposition 4.1}(The Heisenberg distance approximates $d_{CC}$)
Fix $P$ and let $Q \rightarrow P$ such that horizontal coordinates are $O(\epsilon)$ and the missing normal-translation coordinate is $O(\epsilon^2)$. Up to the sign convention for the central Heisenberg coordinate, $(a, \tau)$ defined above are the horizontal norm and weight-two coordinate of the $\mathbb{H}^5$ nilpotentization to leading order. Consequently,
\begin{equation*}
    d_{CC}(P,Q)^2 = H(a^2, w\zeta) + o(\epsilon^2)
\end{equation*}
\begin{proof}
    By rigid motion invariance, we take $P = (0, e_3)$ and choose an orthonormal basis $(e_1, e_2)$ of the tangent frame at $e_3$. By the exponential map, we write the nearby normal as $n_Q = \exp_{e_3}(u)$ with $u\in T_{e_3}\mathbb{S}^2$ and $\|u\|= \alpha$. In this setting, $n_Q = e_3 + u + O(\|u\|^2)$. With $r = r_{T} + r_3e_3$. The symmetric offset $\zeta$ then satisfies
    \begin{equation*}
        \zeta = \left \langle r, \frac{e_3 + n_Q}{2} \right \rangle = r_3 + \frac{1}{2}\inner{r_T}{u} + O(\|r\| \|u\|^2)
    \end{equation*}
    while $\rho^2 = \|r_T\|^2 + o(\epsilon^2)$, $u=O(\epsilon)$ and $\zeta=O(\epsilon^2)$. Rescale the tilt coordinates as $v=wu$. Then, the horizontal metric is Euclidean in $(r_T, v)$ to leading order and has squared norm $\|r_T\|^2 + w^2\|u\|^2=\rho^2 + w^2\alpha^2 + o(\epsilon^2) = a^2 + o(\epsilon^2)$. The contact form is $n \cdot dx$. After the same rescaling, its step-two central coordinate is $w(r_3 + \frac{1}{2}\inner{r_T}{u})$. Hence, $\tau = w\zeta + o(\epsilon^2)$. Mitchell's theorem therefore gives $d_{CC}(P,Q)^2 = H(a^2, w\zeta) + o(\epsilon^2)$. 
\end{proof}


\subsection{Non-asymptotic theoretical guarantees}
\label{app:guarantee}
 
Proposition 4.1 is asymptotic, but we employ the score in more general contexts so we now prove a non-asymptotic property of $\ell_w$ itself.
 
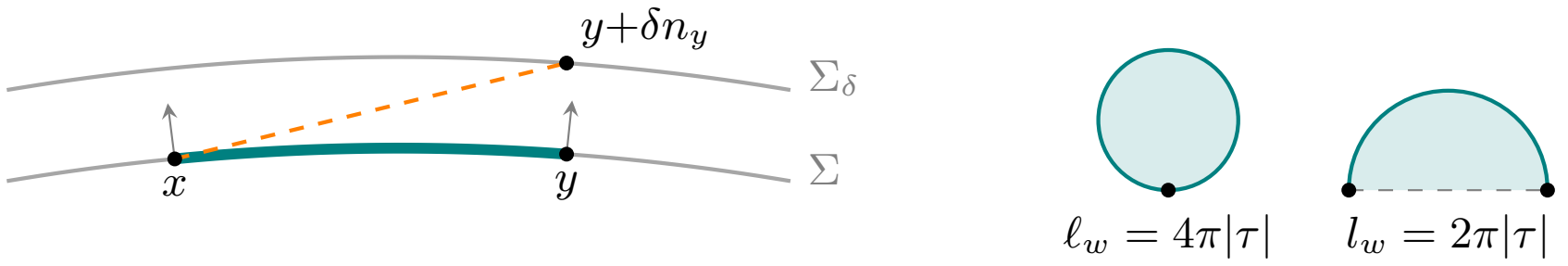
\begin{figure}[ht!]
\centering
\begin{tikzpicture}[x=1cm,y=1cm,>=stealth,font=\small]
  \draw[gray!70,thick] plot[domain=0:5.6,smooth] (\x,{0.30-0.03*(\x-2.8)^2});
  \draw[gray!70,thick] plot[domain=0:5.6,smooth] (\x,{0.95-0.03*(\x-2.8)^2});
  \node[gray,anchor=west] at (5.6,0.80) {$\Sigma_\delta$};
  \node[gray,anchor=west] at (5.6,0.15) {$\Sigma$};
  \draw[teal,line width=2.2pt] plot[domain=1.2:4.0,smooth] (\x,{0.30-0.03*(\x-2.8)^2});
  \draw[orange,thick,dashed] (1.2,0.223) -- (4.0,0.907);
  \draw[->,gray] (1.2,0.223) -- (1.15,0.62);
  \draw[->,gray] (4.0,0.257) -- (4.04,0.64);
  \fill (1.2,0.223) circle (1.5pt) node[below] {$x$};
  \fill (4.0,0.257) circle (1.5pt) node[below] {$y$};
  \fill (4.0,0.907) circle (1.5pt) node[above right=-1pt] {$y{+}\delta n_y$};
  \begin{scope}[xshift=8.3cm]
    \filldraw[teal,fill=teal!15,thick] (0,0.5) circle (0.5);
    \fill (0,0) circle (1.5pt);
    \node[anchor=north] at (0,-0.05) {$\ell_w = 4\pi|\tau|$};
    \filldraw[teal,fill=teal!15,thick] (2.71,0) arc (0:180:0.71);
    \draw[gray,dashed] (1.29,0) -- (2.71,0);
    \fill (1.29,0) circle (1.5pt); \fill (2.71,0) circle (1.5pt);
    \node[anchor=north] at (2.0,-0.05) {$l_w=2\pi|\tau|$};
  \end{scope}
\end{tikzpicture}
\caption{\textbf{Left:} only $\zeta$ sees the sheet, Eq.~\eqref{eq:sheet}. \textbf{Right:} paths enclosing the same area $|\tau|$ and their $\ell^2$: full loop ($a=0$) against an optimal half-loop ($a^2=8|\tau|/\pi$). This is exactly the detour cost illustration.}
\label{fig:guarantee}
\end{figure}

\textbf{The detour cost is an area.} In $\mathbb{H}^5$, a horizontal path gains the central coordinate $\tau$ as the area it sweeps in the (slide, $w\cdot$tilt) plane. So $H(a^2,\tau)$ in equation (6) is the squared length of the shortest arc with chord $a$ enclosing area $|\tau|$ as shown in~\Cref{fig:guarantee} on the right. As mentionned in the main paper in Sec. 4.2, the isoperimetric problem offers lower and upper bounds that we derive now
 
\begin{lemma}[Isoperimetric inequality]\label{lem:dido}
For all $a\ge 0$ and $\tau\in\mathbb{R}$,
$\max\{a^2,\,2\pi|\tau|\}\le H(a^2,\tau)\leq a^2 + 4\pi|\tau|$ and $\sqrt{H(a^2,\tau)}\le a+\sqrt{4\pi|\tau|}$,
and $H(a^2,\tau)=2\pi|\tau|$ if and only if $a^2=8|\tau|/\pi$.
\end{lemma}
\begin{proof}
Upper: slide along the chord (length $a$), then run a full circle of area $|\tau|$ (length $\sqrt{4\pi|\tau|}$). Lower: $a=0$ gives $H=4\pi|\tau|$. Otherwise equation (6) and $\sin\frac{\phi}{2}\le\frac{\phi}{2}$ give $H\ge a^2$, and $H=2|\tau|f(\phi)$ with $f=\phi^2/(\phi-\sin\phi)$. Since $f'$ has the sign of $\cos\frac{\phi}{2}\big(\phi\cos\frac{\phi}{2}-2\sin\frac{\phi}{2}\big)$, negative on $(0,\pi)$ and positive on $(\pi,2\pi)$, $f\ge f(\pi)=\pi$, and $\phi=\pi$ is $s=\pi/2$ in equation (7).
\end{proof}
 
In particular, with $a^2=\rho^2+w^2\alpha^2$, $\tau=w\zeta$, the lemma is a ball-box statement for the score: it is bounded above and below by quantities linear in $|\zeta|$ unlike any quadratic off-plane penalty (\textsc{Ptp}):
\begin{equation}\label{eq:ballbox}
\tfrac12 N^2\le\ell_w^2\le 2N^2,\qquad N^2=\rho^2+w^2\alpha^2+2\pi w|\zeta| .
\end{equation}
The upper bound $a^2 + 4 \pi w |\zeta|$ corresponds to the linear approximation \textsc{LIN}-$4\pi$.
 
\textbf{Stacked sheets.} Let $\Sigma$ be a compact oriented $C^2$ surface with reach $\ge R$ and $\Sigma_\delta=\{x+\delta n_x\}$, $0<\delta<R$, its co-oriented offset. Samples are $P=(x+s\,n_x,n_x)$ and $Q=(y+t\,n_y,n_y)$ with $x,y\in\Sigma$ and $s,t\in\{0,\delta\}$. Write $d=\|x-y\|$ and $\alpha=\angle(n_x,n_y)$. Because $\langle n_x,n_x+n_y\rangle=\langle n_y,n_x+n_y\rangle=2\cos^2\frac{\alpha}{2}$, the symmetric offset equation (5) splits as
\begin{equation}\label{eq:sheet}
\zeta(P,Q)=\zeta_\Sigma(x,y)+(t-s)\cos^2\tfrac{\alpha}{2},\qquad \zeta_\Sigma(x,y)=\tfrac12\langle y-x,\,n_x+n_y\rangle .
\end{equation}
The sheet index enters only through the weight-two coordinate. The reach controls the rest~\citep{federer1959curvature,boissonnat2019reach}: $|\langle y-x,n_x\rangle|\le d^2/2R$ and $\sin\frac{\alpha}{2}\le d/2R$.
 
\begin{theorem}[Separation]\label{thm:sep}
For every $w>0$, we have the following separation properties
\begin{itemize}
    \item (i) if $s=t$ and $d\le R$, then $\ell_w(P,Q)\le C\,d$ with $C=1+\frac{\delta}{R}+\frac{\pi w}{3R}+\sqrt{2\pi w/R}$, and $\ell_w=d$ when $\Sigma$ is a plane;
    \item (ii) if $s\neq t$, then $\ell_w(P,Q)^2\ge\max\{\rho^2+w^2\alpha^2,\ 2\pi w(\delta\cos^2\frac{\alpha}{2}-\frac{d^2}{2R})\}$;
    \item (iii) if $\Sigma$ is a plane, the infimum of $\ell_w^2$ over cross-sheet pairs is $2\pi w\delta$, attained at $\rho^2=8w\delta/\pi$.
\end{itemize}
\end{theorem}
\begin{proof}
Everything comes from the lemma:
(i) $\zeta=\zeta_\Sigma$, so $|\zeta|\le d^2/2R$; also $\rho\le\|r\|\le d+2s\sin\frac{\alpha}{2}\le d(1+\frac{\delta}{R})$ and $\alpha\le\frac{\pi d}{3R}$ and insert into the upper bound of Lemma~\ref{lem:dido}. (ii) \eqref{eq:sheet} with $|t-s|=\delta$ and the lower bound. (iii) $\alpha=0$, $|\zeta|=\delta$: equality case of the lemma.
\end{proof}
 
\begin{corollary}[Leak-free neighborhoods]\label{cor:leak}
Suppose every sample has at least $k$ same-sheet samples within base distance $\varepsilon$. We have the following
\begin{itemize}
    \item[(a)] If $\Sigma$ is a plane and $\varepsilon^2<2\pi w\delta$, every $\ell_w$-kNN set stays on its own sheet.
    \item[(b)] For stacked samples ($x$ and $x+\delta n_x$ both sampled) at same-sheet $k$-th neighbor distance $r_k$, \textsc{Euc}, \textsc{Bil} and \textsc{Prod} leak as soon as $r_k>\delta$, and \textsc{Ptp} as soon as $r_k>\sqrt{1+4\pi}\,\delta$.
\end{itemize}
\end{corollary}

\begin{proof}
\emph{(a)} By Theorem~\ref{thm:sep} (i), same-sheet candidates satisfy $\ell_w\le\varepsilon$. By~\Cref{thm:sep} (iii), cross-sheet candidates satisfy $\ell_w^2\ge 2\pi w\delta>\varepsilon^2$.
 
\emph{(b)} The stacked partner lies at $\|r\|=\delta$ with $\alpha=0$. \textsc{Bil}'s weight is therefore $1$ and \textsc{Prod} adds nothing, so these scores equal $\delta^2$. For \textsc{Ptp}, $\zeta_i=\zeta_j=\delta$ gives $(1+4\pi)\delta^2$. A same-sheet neighbor on a plane has $\zeta_i=\zeta_j=0$, so its score is $r_k^2$.
 
 
 
\emph{Remark (anti-parallel sheets).} If the two faces of a thin leaf carry oriented, opposite normals, then $\alpha\approx\pi$ and~\eqref{eq:ballbox} gives $\ell_w\ge w\alpha\approx\pi w$. They are separated at any thickness, and $\zeta$ plays no role. Under the unoriented score (10), the minimizing branch flips one normal, so the faces become co-oriented and Theorem~\ref{thm:sep} applies as stated.
\end{proof}

\section{Details for the experiments and more results}
\label{app:more_on_the_experiments}

\subsection{Hyperparameters and hardware details}
\label{app:hyper}

\paragraph{Scores.} In all PointNet++ experiments, $h$ is the per-shape median Euclidean $k$-th-neighbor distance, recomputed at each set-abstraction level. Each grouping parameter is optimized with $w=h$ for \textsc{Contact} and \textsc{Prod}. \textsc{Ptp} weights $\zeta_i^2+\zeta_j^2$ by $2\pi$, and \textsc{Bil} uses $\sigma_\alpha=0.4$.
 
\paragraph{PointNet++ (Sec. 5.2).} Single-scale grouping, four set-abstraction levels with $1024$, $256$, $64$ and $32$ centroids, $k=32$, a feature-propagation decoder, and a $2$-way (Pheno4D) or $6890$-way (FAUST) head. Normals are an input feature for every score. PCA normals uses $24$ neighbors. Pheno4D: $16{,}384$ points per cloud, $4{,}096$ drawn per training sample.
 
\paragraph{Heat diffusion (Sec. 5.3).} Kernels use mesh area weights, \citet{Sharp2020ALF} uses its own point-cloud masses. For the bunny, dragon and stacked sheets heat diffusion experiment, we report in~\Cref{tab:heat_exp_details} more details on the parameters.

\begin{table}[ht!]
    \centering
    \begin{tabular}{l|c c c c}
    \toprule
         Shape & $\sigma$ & w & Nb points & Final time $T$\\ \midrule
         Bunny & 0.028 & 0.028 & 2{,}534 & 0.024 \\
         Dragon & 0.013 & 0.013 & 12{,}507 & 0.024 \\
         Stacked sheets & 1/12 & 2/3 & 1{,}875 & 0.12 \\ \bottomrule
    \end{tabular}
    \label{tab:heat_exp_details}
\end{table}
Bunny and Dragon meshes have unit bounding-box diagonal and each sheet has unit side length. Curvature caps the benefit of large $w$ as shown in App.~\ref{subapp:separation_sweep}. Thus, \(w=\sigma\) for the animals and \(w=8\sigma\) for the sheets.
 
\paragraph{Hardware and runtime.} All experiments were conducted on a workstation equipped with a single NVIDIA GeForce RTX 6000 Ada GPU with 32\,GB of VRAM. Training with PointNet++ is done with 100 epochs each, with Adam optimizer and a batch size of 8.

\subsection{The separation in practice: synthetic exploration}
\label{subapp:separation_sweep}

\begin{figure}[ht!]
\centering
\includegraphics[width=\linewidth]{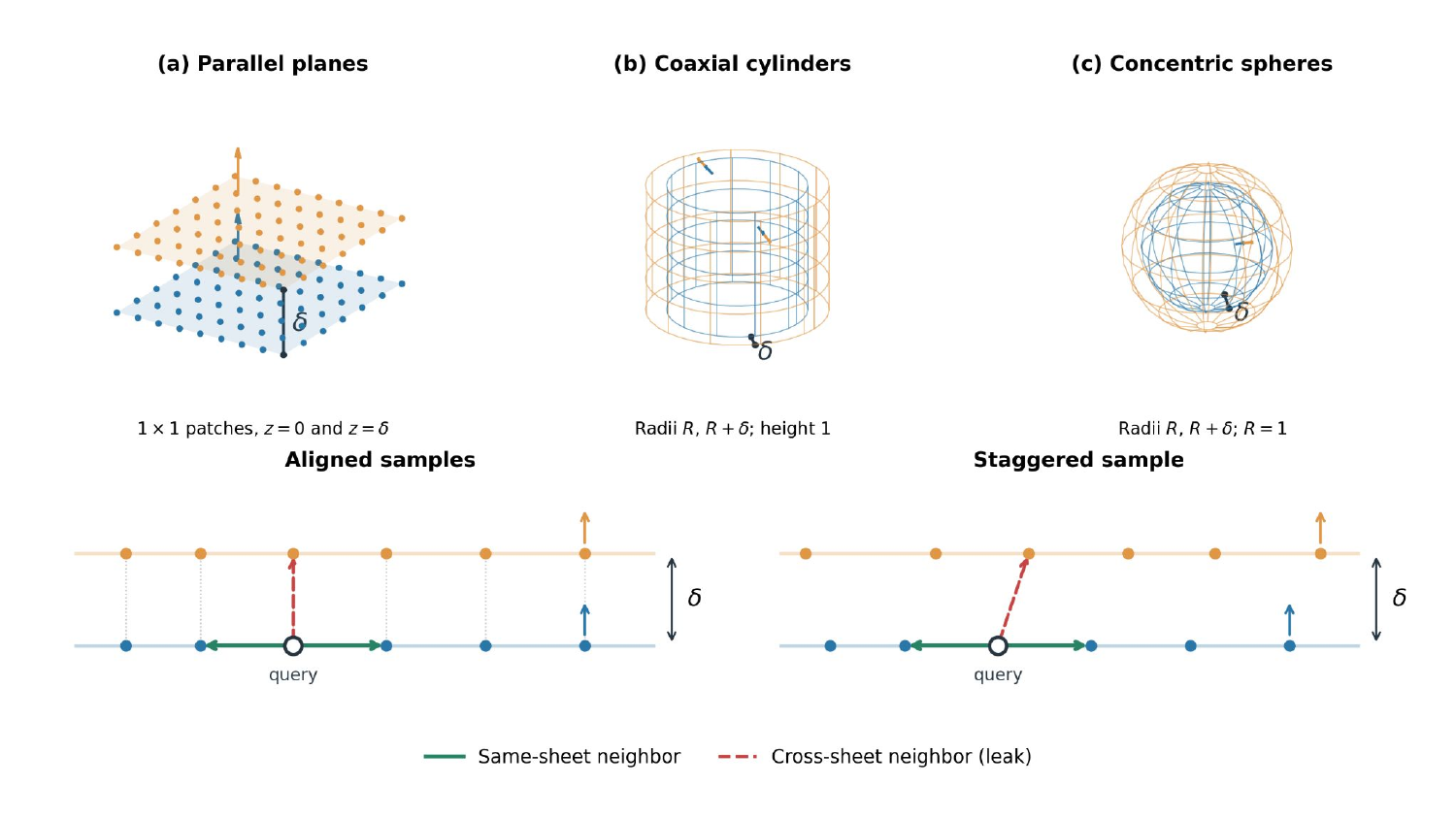}
\vspace{-5mm}
\caption{\textbf{Setting of the experiment and section of the contact-prone regions.} We define two nearby sheets that should aggregate into separate neighborhoods (blue and orange). Normals point in the same local direction. A correct edge stays on its query's sheet while a red edge crosses the gap $\delta$ and count as a leak.}
\label{fig:sradd-setting-brief}
\end{figure}

In this section, we test robustness of the proposed score compared to the baselines. We experiment on sampled synthetic shapes parallel planes, coaxial cylinders and concentric spheres as illustrated in~\Cref{fig:sradd-setting-brief}. Each cloud contains 2,500 points per sheet. Planes have unit side length, cylinders have unit height, and the inner cylinder and sphere have radius $R=1$. Samples follow a Poisson-disk distribution. The two sheets have co-oriented normals before noise is added.

\paragraph{How leakage is measured.}
For every sampled point $i$, we select its $k=16$ nearest neighbors under the tested score. An edge $i\to j$ is incorrect when $i$ and $j$ belong to different sheets. If $s_i$ is the known sheet label and $\mathcal N_k(i)$ the selected neighbors, the reported percentage is
\[
 L=\frac{100}{Nk}\sum_{i=1}^{N}\sum_{j\in\mathcal N_k(i)}
 \mathbf{1}\{s_i\ne s_j\}.
\]
$L=0\%$ means perfect separation, and $L=50\%$ means half of the selected edges cross the gap. Each run evaluates all $N=5{,}000$ queries and 80,000 directed edges. We average over ten independent samplings and~\Cref{tab:sradd-leakage-brief} reports the sample standard deviation across these samplings. CONTACT uses $H(\rho^2+w^2\alpha^2,w\zeta)$; LIN adds $2\pi w|\zeta|$
to $\rho^2+w^2\alpha^2$, and QUAD adds $4\pi\zeta^2$. BIL reweights an EUC graph, unlike the angular grouping baseline in the archived learning runs.


\begin{table}
\centering
\small
\setlength{\tabcolsep}{3.5pt}
\renewcommand{\arraystretch}{1.03}
\begin{tabular}{llcccc}
\toprule
Geometry & Grouping & Base & Staggered & Base+noise & Staggered+noise \\
 & & $0^\circ$ & $0^\circ$ & $5^\circ$ & $5^\circ$ \\
\midrule
Planes & EUC / BIL & $49.17\,{\scriptstyle\pm}\,0.06$ & $50.46\,{\scriptstyle\pm}\,0.09$ & $49.17\,{\scriptstyle\pm}\,0.06$ & $50.46\,{\scriptstyle\pm}\,0.09$ \\
 & PROD (ang.) & $49.17\,{\scriptstyle\pm}\,0.06$ & $50.46\,{\scriptstyle\pm}\,0.09$ & $50.69\,{\scriptstyle\pm}\,0.08$ & $50.27\,{\scriptstyle\pm}\,0.06$ \\
 & PTP & $23.15\,{\scriptstyle\pm}\,0.13$ & $22.91\,{\scriptstyle\pm}\,0.10$ & $25.50\,{\scriptstyle\pm}\,0.14$ & $25.36\,{\scriptstyle\pm}\,0.15$ \\
 & LIN & $0.00\,{\scriptstyle\pm}\,0.00$ & $0.00\,{\scriptstyle\pm}\,0.00$ & $1.33\,{\scriptstyle\pm}\,0.10$ & $1.30\,{\scriptstyle\pm}\,0.10$ \\
 & QUAD & $25.45\,{\scriptstyle\pm}\,0.13$ & $25.06\,{\scriptstyle\pm}\,0.09$ & $32.86\,{\scriptstyle\pm}\,0.26$ & $32.59\,{\scriptstyle\pm}\,0.18$ \\
 & CONTACT & $0.00\,{\scriptstyle\pm}\,0.00$ & $0.00\,{\scriptstyle\pm}\,0.00$ & $1.18\,{\scriptstyle\pm}\,0.10$ & $1.16\,{\scriptstyle\pm}\,0.11$ \\
\midrule
Cylinders & EUC / BIL & $49.18\,{\scriptstyle\pm}\,0.03$ & $50.47\,{\scriptstyle\pm}\,0.11$ & $49.18\,{\scriptstyle\pm}\,0.03$ & $50.47\,{\scriptstyle\pm}\,0.11$ \\
 & PROD (ang.) & $49.55\,{\scriptstyle\pm}\,0.02$ & $51.06\,{\scriptstyle\pm}\,0.06$ & $51.08\,{\scriptstyle\pm}\,0.12$ & $50.69\,{\scriptstyle\pm}\,0.12$ \\
 & PTP & $24.18\,{\scriptstyle\pm}\,0.15$ & $23.88\,{\scriptstyle\pm}\,0.09$ & $26.33\,{\scriptstyle\pm}\,0.13$ & $26.15\,{\scriptstyle\pm}\,0.10$ \\
 & LIN & $0.00\,{\scriptstyle\pm}\,0.00$ & $0.00\,{\scriptstyle\pm}\,0.00$ & $1.17\,{\scriptstyle\pm}\,0.07$ & $1.18\,{\scriptstyle\pm}\,0.10$ \\
 & QUAD & $33.44\,{\scriptstyle\pm}\,0.10$ & $32.89\,{\scriptstyle\pm}\,0.10$ & $36.40\,{\scriptstyle\pm}\,0.13$ & $36.15\,{\scriptstyle\pm}\,0.11$ \\
 & CONTACT & $0.00\,{\scriptstyle\pm}\,0.00$ & $0.00\,{\scriptstyle\pm}\,0.00$ & $1.13\,{\scriptstyle\pm}\,0.09$ & $1.15\,{\scriptstyle\pm}\,0.10$ \\
\midrule
Spheres & EUC / BIL & $49.14\,{\scriptstyle\pm}\,0.02$ & $50.54\,{\scriptstyle\pm}\,0.07$ & $49.14\,{\scriptstyle\pm}\,0.02$ & $50.54\,{\scriptstyle\pm}\,0.07$ \\
 & PROD (ang.) & $49.85\,{\scriptstyle\pm}\,0.01$ & $51.88\,{\scriptstyle\pm}\,0.06$ & $51.76\,{\scriptstyle\pm}\,0.14$ & $51.37\,{\scriptstyle\pm}\,0.12$ \\
 & PTP & $26.02\,{\scriptstyle\pm}\,0.15$ & $25.52\,{\scriptstyle\pm}\,0.08$ & $27.75\,{\scriptstyle\pm}\,0.14$ & $27.45\,{\scriptstyle\pm}\,0.09$ \\
 & LIN & $0.00\,{\scriptstyle\pm}\,0.00$ & $0.00\,{\scriptstyle\pm}\,0.00$ & $1.03\,{\scriptstyle\pm}\,0.11$ & $1.03\,{\scriptstyle\pm}\,0.12$ \\
 & QUAD & $42.33\,{\scriptstyle\pm}\,0.09$ & $42.33\,{\scriptstyle\pm}\,0.07$ & $42.64\,{\scriptstyle\pm}\,0.19$ & $42.36\,{\scriptstyle\pm}\,0.17$ \\
 & CONTACT & $0.00\,{\scriptstyle\pm}\,0.00$ & $0.00\,{\scriptstyle\pm}\,0.00$ & $1.16\,{\scriptstyle\pm}\,0.11$ & $1.16\,{\scriptstyle\pm}\,0.12$ \\
\bottomrule
\end{tabular}
\caption{\textbf{Leakage in the four configurations.}
Cross-sheet edges in percent, mean $\pm$ sample stand. deviations over ten samplings and lower is better.}
\label{tab:sradd-leakage-brief}
\end{table}

\begin{figure}
\begin{minipage}{\linewidth}
\centering
\includegraphics[width=\linewidth]{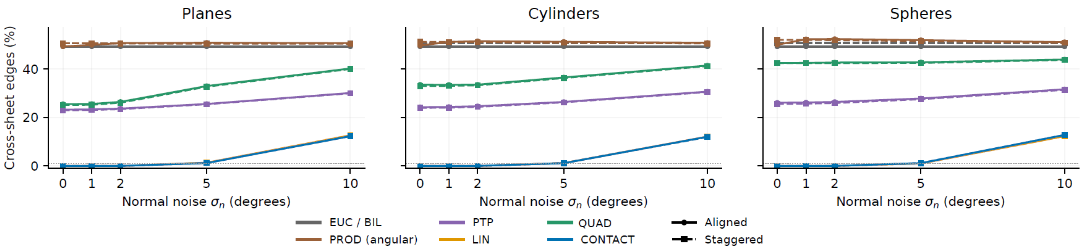}
\caption{\textbf{Leakage versus normal noise at the same fixed gap.} Lines show ten-sampling means. Solid/circle curves use aligned samples; dashed/square curves use staggered samples. The dotted line marks $1\%$ leakage. PROD denotes the angular product in this synthetic experiment.}
\label{fig:sradd-noise-leakage}
\end{minipage}
\end{figure}

\paragraph{Analysis.}
With exact normals, CONTACT and LIN have zero leakage in every tested geometry and both sampling schemes at this operating point. Under $5$° noise, their leakage is approximately $1.0$-$1.3\%$, compared with $23$-$28\%$ for PTP and roughly $49$-$52\%$ for EUC/BIL and PROD. Staggering has little effect on this ordering, whereas increasing normal noise has a visible effect on the contact score, especially for more than 5° noise. CONTACT and LIN behave similarly.


\subsection{More qualitative results}
\subsubsection{Plant segmentation}
\begin{figure}[ht!]
  \centering
  \includegraphics[trim={2.4cm 0 2.4cm 0},clip,width=1.0\linewidth]{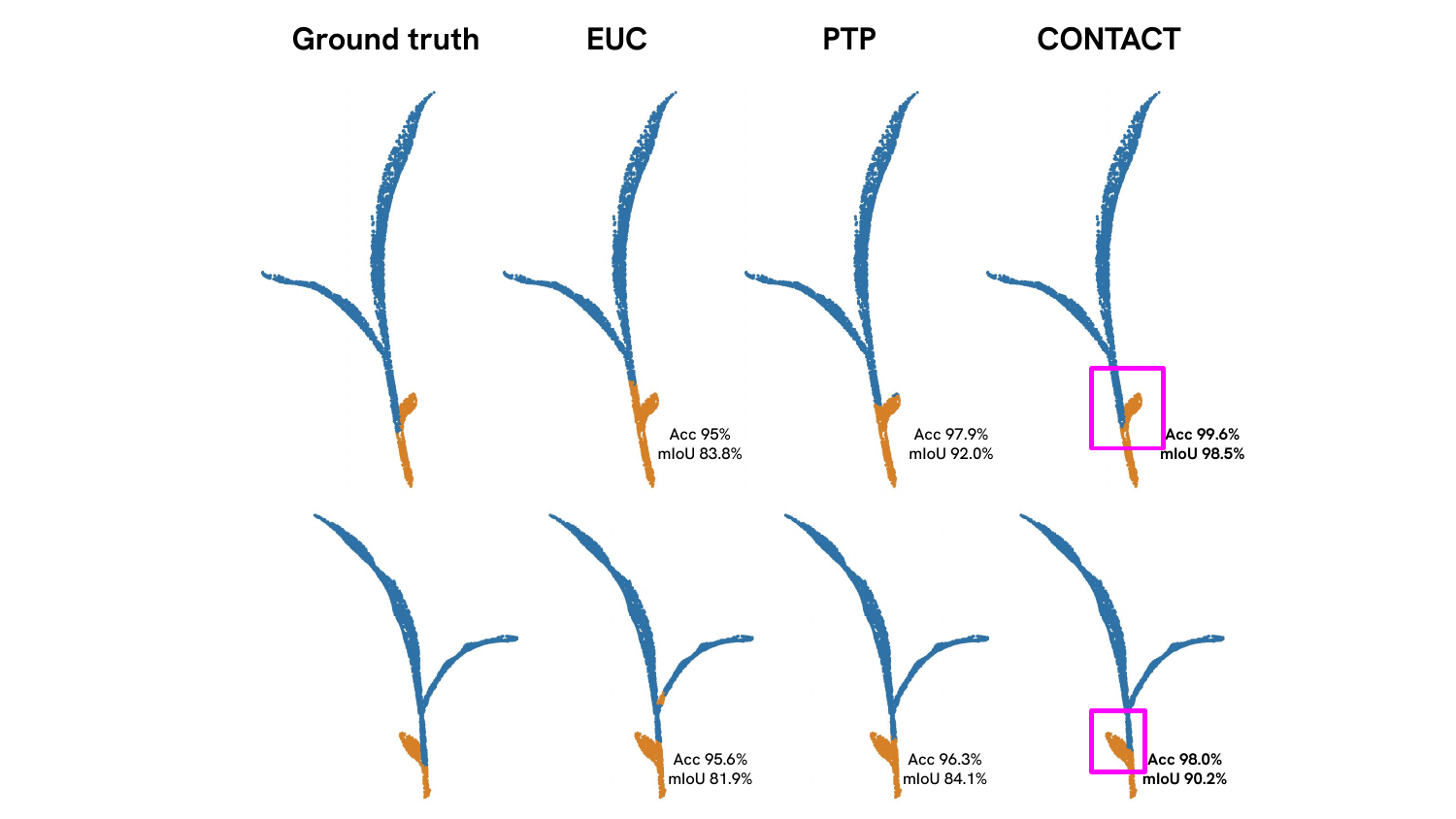}
  \caption{\textbf{Qualitative visualization for the segmentation experiment on plant point clouds.} .}
  \label{fig:more_qual_pheno4D}
\end{figure}

Here, we report qualitative comparisons of the predicted segmentation maps compared to the ground truth in~\Cref{fig:more_qual_pheno4D}.

\subsubsection{Human body correspondence}
\begin{figure}[ht!]
  \centering
  \includegraphics[width=\linewidth]{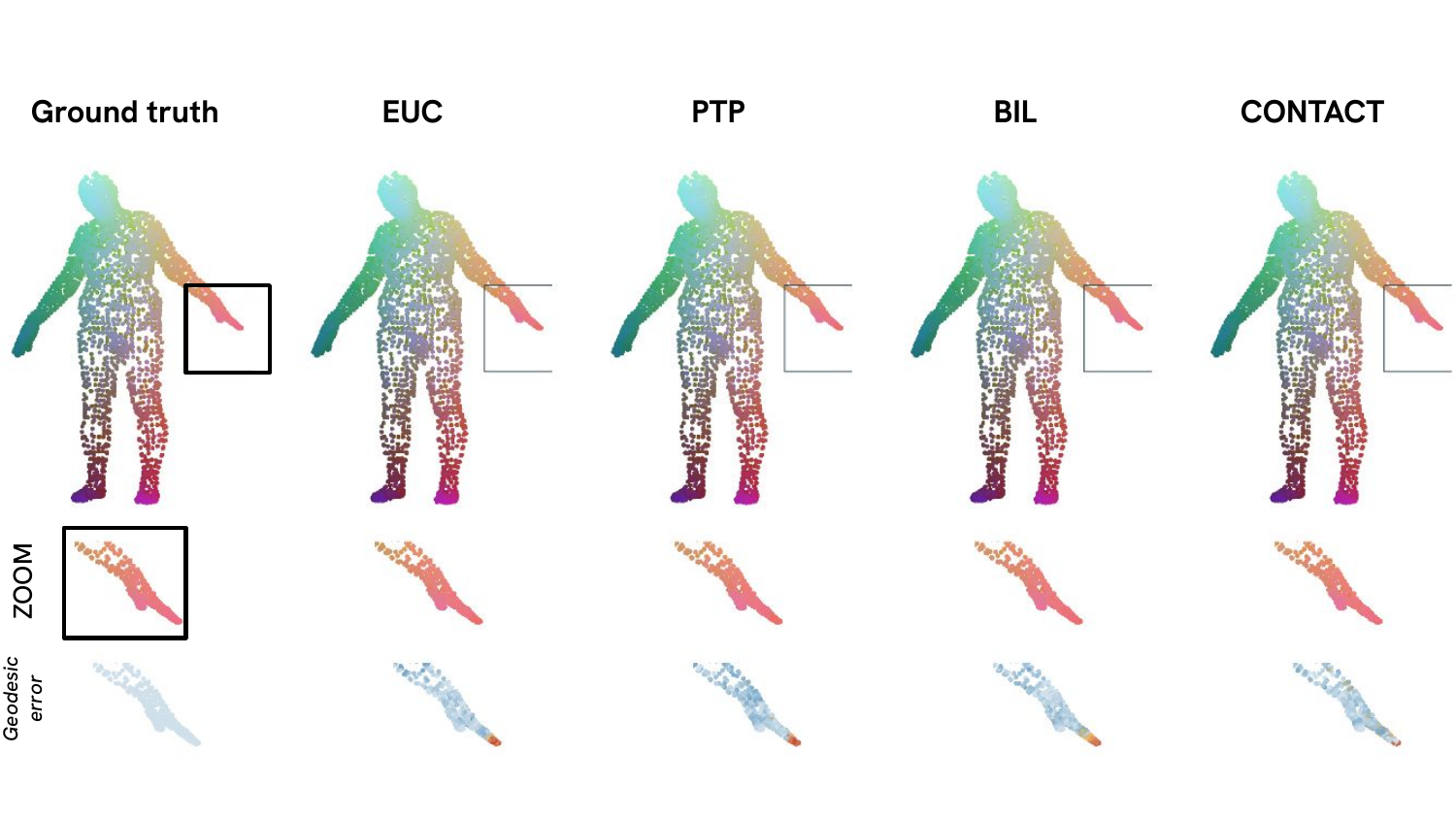}
  \caption{\textbf{Qualitative visualization for the correspondence experiment on human body meshes.} We show a qualitative example of predicted correspondence map on a test subject. Zooming on the left hand to highlight the effect of using the CONTACT neighborhood in a PointNet++ architecture.}
  \label{fig:human_corres_qual}
\end{figure}

We report a qualitative example of result for the correspondence experiment in~\Cref{fig:human_corres_qual}. A zoom on a contact-prone region (around the fingers) highlight the effect of the CONTACT score.

\subsubsection{PDE on oriented point clouds}

\begin{figure}[ht!]
  \centering
  \includegraphics[width=\linewidth]{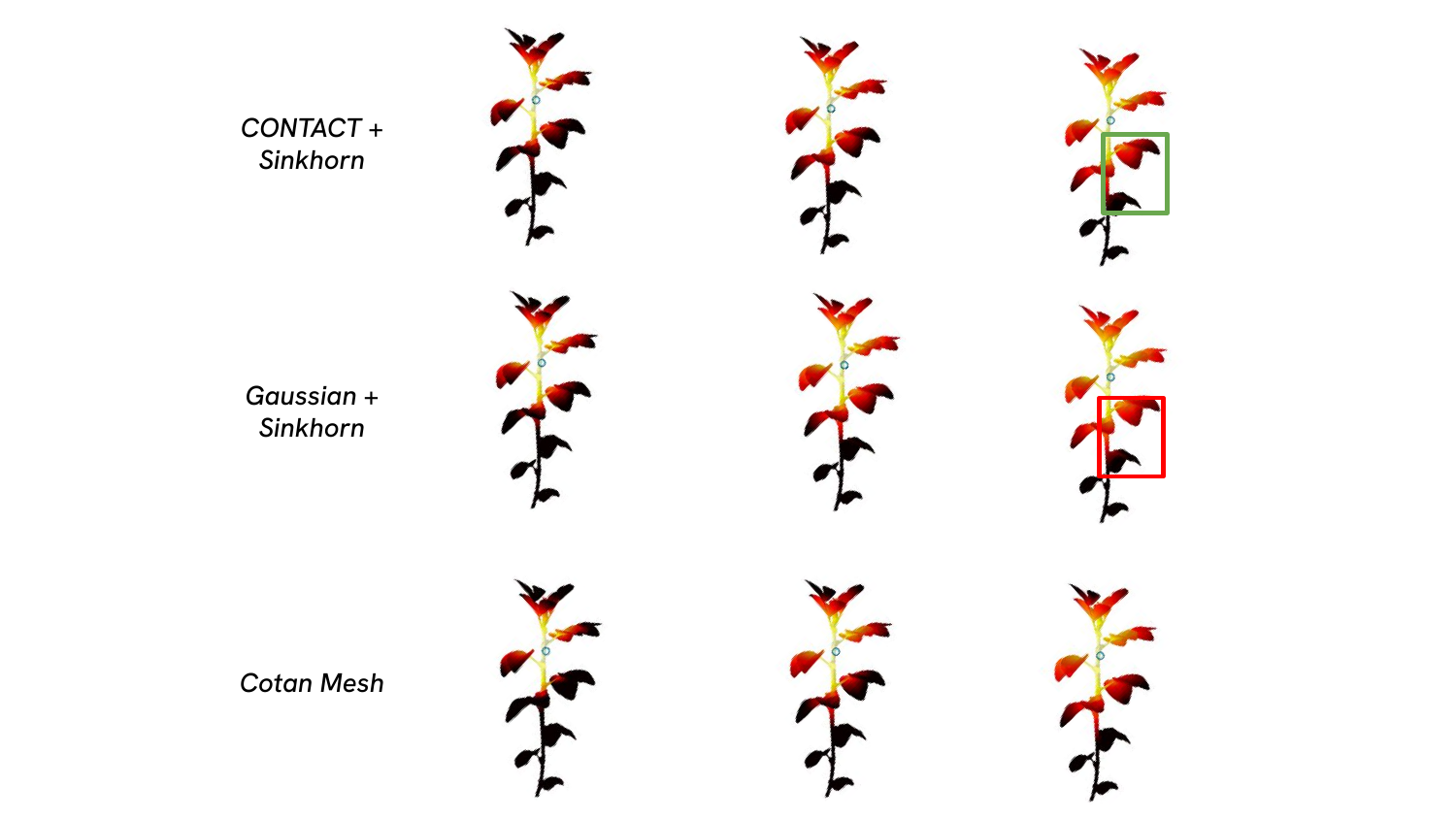}
  \caption{\textbf{Additional qualitative comparison of heat diffusion.} We report a visual example for plant point cloud obtained with a mesh reconstruction algorithm \citep{gomez2026blobsspokeshighfidelitysurface} from optimized gaussian splatting \citep{kerbl3Dgaussians}. In particular, we highlight how the original Gaussian similarity induces more shortcut leakage than with the CONTACT similarity.}
  \label{fig:heat_qual_faba}
\end{figure}

We repeat the heat experiment of Sec. 5.3 on one of our own plant reconstruction using Gaussian splatting \cite{kerbl3Dgaussians} and report a qualitative visualization in~\Cref{fig:heat_qual_faba}.

\begin{figure}[ht!]
  \centering
  \includegraphics[width=\linewidth]{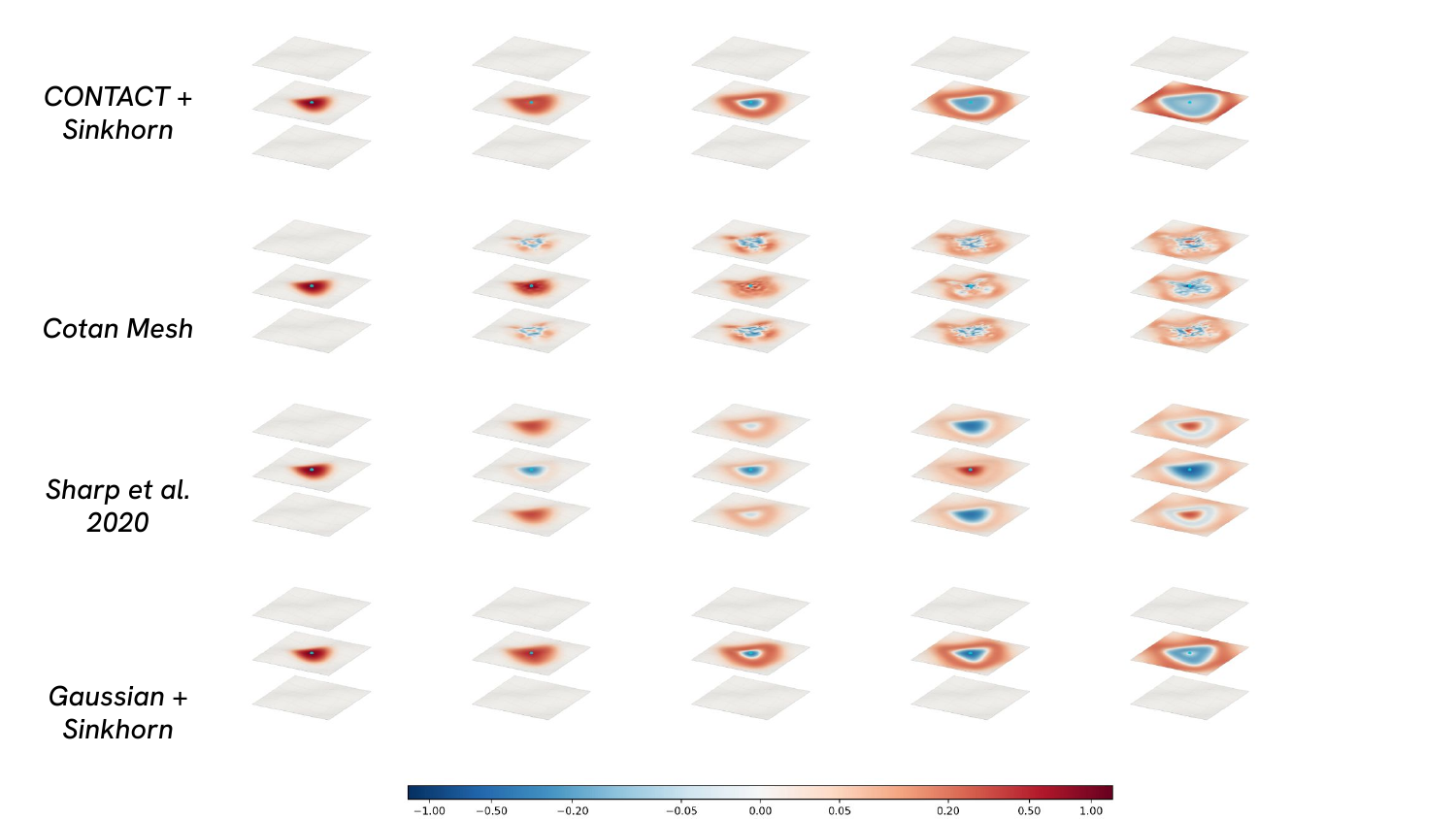}
  \caption{\textbf{A qualitative example of wave propagation using a similar approach.} Three stacked sheets, similar than the ones from Sec. 5.3 with a wave pulse at the center of the middle sheet. We highlight how, for Gaussian-Sinkhorn and point cloud laplacian, propagation is quickly corrupted by the other sheets where CONTACT-Sinkhorn stays closer to the reference mesh-based propagation.}
  \label{fig:wave_prop_qual_sheets}
\end{figure}

We go a little further and define wave propagation using a similar approach. 
\paragraph{Wave propagation on wavy sheets.}
We sample three surfaces$z=A\sin(2\pi x)\sin(2\pi y)+j/30$, with $(x,y)\in[-\tfrac12,\tfrac12]^2$, $j\in\{-1,0,1\}$,
and $A\in\{0.04,0.16\}$, using $31\times31$ vertices per sheet. The lower two sheets are co-oriented; the upper sheet has reversed normals. A Gaussian displacement of width $0.1$ initializes the middle sheet,
with zero initial velocity. We solve $M\ddot u+Lu=0$ by full modal decomposition, comparing the cotangent mesh reference, the point-cloud Laplacian from \cite{Sharp2020ALF}, Gaussian-Sinkhorn and CONTACT-Sinkhorn operators.
For both kernels, $\sigma=2/30$ and$L=M(I-Q)/\tau$, where $Q$ is the balanced operator, $\tau=\sigma^2/2$, and CONTACT uses $w=8\sigma$. We display five equally spaced times over $t\in[0,0.42]$, preserving signed wave amplitudes without damping or clamping. We report qualitative comparisons in~\Cref{fig:wave_prop_qual_sheets}.

\section{Additional analyses and controls}
Unless stated otherwise, the tables below use the trained networks' first-level neighborhoods: identical point samples, 1,024 farthest-point queries and $k=32$ for every score, each score with the parameters of its learning run. LIN-$4\pi$ is the coefficient-matched control of Section~\ref{sec:experiments}.

\paragraph{Cross-leaf connections on Pheno4D.} On all 36 test scans, we report the share of points a neighborhood method keeps while they should not because it belongs to another At 4,096 points, the neighbor sets are exactly those behind Table~\ref{tab:pheno4d}. The "kept" row is the share of kept neighborhood point that are also in the euclidean neighborhood.
\begin{center}\small
\begin{tabular}{lcccccccc}
\toprule
 & \textsc{Euc} & \textsc{Bil} & \textsc{Prod} & \textsc{Ptp} & QUAD & LIN-$2\pi$ & LIN-$4\pi$ & \textsc{Contact}\\
\midrule
cross-leaf rate & 0.281 & 0.253 & 0.255 & 0.173 & 0.169 & 0.154 & 0.149 & 0.158\\
kept & 1 & 0.87 & 0.88 & 0.82 & 0.65 & 0.65 & 0.63  & 0.64\\
\bottomrule
\end{tabular}
\end{center}
Across densities and proximity thresholds (number of queries in parentheses):
\begin{center}\small
\begin{tabular}{llcccccc}
\toprule
points & other leaf within & \textsc{Euc} & \textsc{Ptp} & QUAD & LIN-$2\pi$ & LIN-$4\pi$ & \textsc{Contact}\\
\midrule
4,096 & $h$ (1,039) & 0.281 & 0.173 & 0.169 & 0.154 & 0.149 & 0.158\\
8,192 & $h$ (592) & 0.298 & 0.192 & 0.192 & 0.178 & 0.176 & 0.181\\
16,384 & $h$ (274) & 0.292 & 0.223 & 0.253 & 0.230 & 0.224 & 0.233\\
4,096 & $h/2$ (187) & 0.431 & 0.383 & 0.364 & 0.349 & 0.341 & 0.352\\
16,384 & $h/2$ (52) & 0.424 & 0.395 & 0.424 & 0.401 & 0.390 & 0.410\\
\bottomrule
\end{tabular}
\end{center}
Scores that charge $\zeta$ remove about 44\% of cross-leaf connections at the training density. The reduction shrinks as leaves come closer relative to the sampling. Touching leaves explain the closest cases: in a separate analysis of touching leaf pairs at 40,000 points, 200,000 points and native resolution (about one million points), the closest gap stayed at $0.08$--$0.12\,h$. It shrinks with the sampling, so it is never resolved


\paragraph{Matched controls on the cylinders.} Same 24 points, normals, $w=R$ and two neighbors as Figure~\ref{fig:synthetic-reconstruction}. Edges and components refer to the undirected union of the neighbor graphs.
\begin{center}\small
\resizebox{\linewidth}{!}{\begin{tabular}{lccccccc}
\toprule
 & \textsc{Euc} & \textsc{Prod} (ang.) & \textsc{Ptp} & QUAD-$4\pi$ & LIN-$2\pi$ & LIN-$4\pi$ & \textsc{Contact}\\
\midrule
first neighbor on the correct cylinder & 0\% & 0\% & 0\% & 0\% & 0\% & 100\% & 100\%\\
two-neighbor purity & 25\% & 25\% & 50\% & 50\% & 50\% & 100\% & 100\%\\
cross-cylinder edges & 24 & 24 & 12 & 12 & 12 & 0 & 0\\
components & 4 & 3 & 4 & 5 & 4 & 2 & 2\\
\bottomrule
\end{tabular}}
\end{center}
Tuning the coefficient, LIN succeeds for $\lambda>7.33\approx2.33\pi$.

\paragraph{Diffusion: kernels, evolution rule, start, sources and $w$.} Each score is swapped into the kernel of Section~\ref{sec:oriented-diffusion} with everything else as for \textsc{Contact} described in Sec. 5.3. Final discrepancy is in \%:
\begin{center}\small
\begin{tabular}{lcccccccc}
\toprule
 & \textsc{Contact} & LIN-$2\pi$ & LIN-$4\pi$ & QUAD-$4\pi$ & \textsc{Prod} (ang.) & \textsc{Ptp} & Gaussian\\
\midrule
sheets & 0.30 & 0.29 & \textbf{0.28} & 50.00 & 50.00 & 66.67 & 66.67\\
bunny & \textbf{2.43} & 5.40 & 10.12 & 3.81 & 3.10 & 9.15 & 7.56\\
dragon & 3.92 & 8.73 & 13.68 & 7.63 & \textbf{3.45} & 15.48 & 10.45\\
\bottomrule
\end{tabular}
\end{center}
A linear charge penalizes the small offsets that same-surface pairs have on curved, sampled surfaces at full slope, whereas $\ell_w^2$ adds only $12w^2\zeta^2/a^2$ there. The larger the coefficient, the worse. On the curved shapes most of the gain of \textsc{Contact} over the Gaussian kernel comes from the angular term. Both kernels under both evolution rules, with positions, normals, bandwidth, masses and source fixed ($s=\sigma^2/2$):
\begin{center}\small
\resizebox{\linewidth}{!}{\begin{tabular}{llcccc}
\toprule
shape & time & \textsc{Contact}, continuous & \textsc{Contact}, fixed steps & Gaussian, continuous & Gaussian, fixed steps\\
\midrule
bunny & $s$ & 41.72 & 26.70 & 41.56 & 5.44\\
 & $8s$ & 9.18 & 7.10 & 4.84 & 3.17\\
 & final & 2.43 & 2.67 & 7.56 & 7.83\\
dragon & $s$ & 48.99 & 25.83 & 49.61 & 6.09\\
 & $8s$ & 5.18 & 2.46 & 7.26 & 8.89\\
 & final & 3.92 & 3.86 & 10.45 & 10.49\\
sheets & $s$ & 38.15 & 2.15 & 75.49 & 66.21\\
 & $8s$ & 3.16 & 0.45 & 66.64 & 66.67\\
 & final & 0.30 & 0.08 & 66.67 & 66.67\\
\bottomrule
\end{tabular}}
\end{center}
The generator's zero-step term retains at least a fraction $e^{-t/s}$ of the heat at the source, which explains most of the early spike under the continuous rule for both kernels. Starting instead from a shared smooth field, under fixed steps:
\begin{center}\small
\begin{tabular}{llcccc}
\toprule
shape & kernel & at $s$ & peak & time average & final\\
\midrule
bunny & \textsc{Contact} & 4.92 & 5.09 & 3.10 & 2.84\\
 & Gaussian & 2.98 & 8.02 & 4.93 & 8.02\\
dragon & \textsc{Contact} & 4.21 & 4.21 & 3.43 & 3.85\\
 & Gaussian & 3.39 & 10.51 & 8.53 & 10.51\\
sheets & \textsc{Contact} & 0.54 & 0.54 & 0.25 & 0.07\\
 & Gaussian & 66.21 & 66.67 & 64.63 & 66.67\\
\bottomrule
\end{tabular}
\end{center}
Under the continuous rule the picture is the same. 
We report results when tuning $w$ with respect to $\sigma$, mean final discrepancy over the five sources against $w$ (point source, continuous rule):
\begin{center}\small
\begin{tabular}{lcccccc}
\toprule
$w/\sigma$ & 0.5 & 1 & 2 & 4 & 8 & 16\\
\midrule
bunny (Gaussian: 6.13) & 4.78 & 2.58 & 11.44 & 37.72 & 78.94 & --\\
dragon (Gaussian: 10.55) & 7.87 & 2.69 & 13.73 & 38.87 & 76.96 & --\\
sheets (Gaussian: 66.67) & -- & 56.02 & 49.81 & 14.41 & 0.44 & 0.42\\
\bottomrule
\end{tabular}
\end{center}
No single $w$ serves both kinds of shape: at large $w$, the turning of the normal along a curved surface is priced like a gap (the factor $1+w^2\kappa^2$ below), and heat stays near the source. The angular product behaves the same way on the curved shapes and never drops below 50\% on the sheets.

\paragraph{Same-surface neighbors lost as $w$ grows.} The two parts of the score react differently to curvature. The second-order terms of $\zeta_P$ and $\zeta_Q$ cancel, so $\zeta=O(\|r\|^3)$ on a smooth surface and $\zeta=0$ exactly on spheres and cylinders. The angular term does not: along a direction of normal curvature $\kappa=1/R$, $\alpha\approx\kappa r$ and $a^2\approx r^2(1+w^2\kappa^2)$, a factor of 6\% at $w=R/4$ and 2 at $w=R$. On a sphere the factor is isotropic and the kNN set does not change; on a cylinder or a saddle the patch shrinks across the curvature and, $k$ being fixed, stretches along the flatter direction. Over all first-level queries at 4,096 points, retention is the share of \textsc{Euc}'s same-surface neighbors still selected, and ``most curved'' is the quarter of queries whose Euclidean neighbors' normals turn most:
\begin{center}\small
\begin{tabular}{lccccc}
\toprule
$w/h$ & 0.5 & 1 & 2 & 4 & 8\\
\midrule
Pheno4D, \textsc{Contact}, retention, all queries & 0.98 & 0.94 & 0.88 & 0.78 & 0.67\\
Pheno4D, \textsc{Contact}, retention, most curved & 0.94 & 0.87 & 0.75 & 0.61 & 0.48\\
Pheno4D, \textsc{Prod}, retention, most curved & 0.98 & 0.91 & 0.78 & 0.61 & 0.45\\
Pheno4D, \textsc{Contact}, cross-leaf rate  & 0.235 & 0.193 & 0.167 & 0.158 & 0.165\\
FAUST, \textsc{Contact}, retention, all queries & 0.98 & 0.94 & 0.87 & 0.77 & 0.66\\
FAUST, \textsc{Contact}, retention, most curved & 0.95 & 0.88 & 0.77 & 0.63 & 0.50\\
FAUST, \textsc{Prod}, retention, most curved & 0.97 & 0.91 & 0.79 & 0.63 & 0.47\\
FAUST, \textsc{Contact}, short-circuit rate  & 0.172 & 0.134 & 0.067 & 0.050 & 0.058\\
\bottomrule
\end{tabular}
\end{center}
The loss comes from the angular term: the angular product loses the same neighbors at the same $w$ while barely reducing cross-surface edges. Beyond $w=2h$ to $4h$ separation no longer improves while retention keeps falling. FAUST with PCA normals gives the same picture.


\paragraph{Sensitivity on real data.} One factor is changed at a time from the training setting ($k=32$ and PCA normals from 24 neighbors on Pheno4D and 16 on FAUST and 4,096 points), with $w$ fixed in absolute units. Accuracy is from single runs of 150 epochs. The first row is rerun with this grid. ``Removed'' is the share of cross-surface edges removed relative to \textsc{Euc}, with the protocols above. E / C is \textsc{Euc} / \textsc{Contact}.
\begin{center}\small
\begin{tabular}{lcccc}
\toprule
 & \multicolumn{2}{c}{Pheno4D} & \multicolumn{2}{c}{FAUST, PCA normals}\\
setting & removed & mIoU, E / C & removed & contact stratum, E / C\\
\midrule
training setting & 44\% & 0.808 / 0.837 & 39\% & 0.077 / 0.049\\
$k=16$ & 40\% & 0.762 / 0.671 & 46\% & 0.089 / 0.089\\
$k=64$ & 38\% & 0.827 / 0.816 & 24\% & 0.069 / 0.048\\
PCA from 12 / 8 neighbors & 49\% & 0.801 / 0.796 & 48\% & 0.093 / 0.070\\
PCA from 48 / 32 neighbors & 35\% & 0.841 / 0.823 & 27\% & 0.099 / 0.064\\
2,048 points & -- & 0.735 / 0.828 & -- & 0.047 / 0.027\\
8,192 points & 39\% & 0.738 / 0.751 & -- & 0.134 / 0.123\\
\bottomrule
\end{tabular}
\end{center}
Generally, a more precise sampling yields better accuracy using the proposed contact score.

\paragraph{Choosing $w$.} $w$ is a length: reorienting by $\alpha$ costs the same as sliding $w\alpha$. With $\varepsilon$ the same-surface $k$-th neighbor distance, $\delta$ the gap, $\sigma_n$ the normal noise and $R$ the curvature radius, a heuristic window is: $w>\varepsilon^2/(2\pi\delta)$ for separation on planes (Corollary~\ref{cor:leak}); $w$ near $0.3\,\varepsilon/\sigma_n$ under noise, gaps below about $2.1\,\varepsilon\sigma_n$ being inseparable; and, for curvature, same-surface costs grow by $1+w^2\kappa^2$ (on spheres, separation degrades between $w\approx1.3R$ and $5.2R$). On real data we recommend the knee at $w\approx2h$ to $4h$, chosen for the $k$ and density in use. 



%% file: references.bib
@inproceedings{bekkers2024ponita,
  title={Fast, Expressive {SE}$(n)$ Equivariant Networks through Weight-Sharing in Position-Orientation Space},
  author={Bekkers, Erik J and Vadgama, Sharvaree and Hesselink, Rob and van der Linden, Putri A and Romero, David W},
  booktitle={International Conference on Learning Representations (ICLR)},
  year={2024}
}

@article{duits2016subriemannian,
  title={Sub-{R}iemannian geodesics in {$SE(3)$} whose spatial projections do not have cusps},
  author={Duits, Remco and Ghosh, Arpan and Dela Haije, Tom and Mashtakov, Alexey},
  journal={Journal of Dynamical and Control Systems},
  volume={22},
  number={4},
  pages={771--805},
  year={2016}
}

@article{duits2018reedsshepp,
  title={Optimal paths for variants of the {2D} and {3D} {R}eeds-{S}hepp car with applications in image analysis},
  author={Duits, Remco and Meesters, Stephan PL and Mirebeau, Jean-Marie and Portegies, Jorg M},
  journal={Journal of Mathematical Imaging and Vision},
  volume={60},
  number={6},
  pages={816--848},
  year={2018}
}

@article{franken2009crossing,
  title={Crossing-preserving coherence-enhancing diffusion on invertible orientation scores},
  author={Franken, Erik and Duits, Remco},
  journal={International Journal of Computer Vision},
  volume={85},
  number={3},
  pages={253--278},
  year={2009}
}

@article{smets2023pde,
  title={{PDE}-based group equivariant convolutional neural networks},
  author={Smets, Bart MN and Portegies, Jim and Bekkers, Erik J and Duits, Remco},
  journal={Journal of Mathematical Imaging and Vision},
  volume={65},
  number={1},
  pages={209--239},
  year={2023}
}

@article{bellaard2023analysis,
  title={Analysis of (sub-){R}iemannian {PDE-G-CNN}s},
  author={Bellaard, Gijs and Bon, Daan LJ and Pai, Gautam and Smets, Bart MN and Duits, Remco},
  journal={Journal of Mathematical Imaging and Vision},
  volume={65},
  number={6},
  pages={819--843},
  year={2023}
}

@article{citti2006cortical,
  title={A cortical based model of perceptual completion in the roto-translation space},
  author={Citti, Giovanna and Sarti, Alessandro},
  journal={Journal of Mathematical Imaging and Vision},
  volume={24},
  number={3},
  pages={307--326},
  year={2006}
}

@article{petitot2003neurogeometry,
  title={The neurogeometry of pinwheels as a sub-{R}iemannian contact structure},
  author={Petitot, Jean},
  journal={Journal of Physiology-Paris},
  volume={97},
  number={2-3},
  pages={265--309},
  year={2003}
}

@book{montgomery2002tour,
  title={A Tour of Subriemannian Geometries, Their Geodesics and Applications},
  author={Montgomery, Richard},
  publisher={American Mathematical Society},
  series={Mathematical Surveys and Monographs},
  volume={91},
  year={2002}
}

@article{chow1939systeme,
  title={{\"U}ber {S}ysteme von linearen partiellen {D}ifferentialgleichungen erster {O}rdnung},
  author={Chow, Wei-Liang},
  journal={Mathematische Annalen},
  volume={117},
  pages={98--105},
  year={1939}
}

@inproceedings{qi2017pointnetpp,
  title={Point{N}et++: Deep hierarchical feature learning on point sets in a metric space},
  author={Qi, Charles Ruizhongtai and Yi, Li and Su, Hao and Guibas, Leonidas J},
  booktitle={Advances in Neural Information Processing Systems (NeurIPS)},
  year={2017}
}

@article{wang2019dynamic,
  title={Dynamic graph {CNN} for learning on point clouds},
  author={Wang, Yue and Sun, Yongbin and Liu, Ziwei and Sarma, Sanjay E and Bronstein, Michael M and Solomon, Justin M},
  journal={ACM Transactions on Graphics},
  volume={38},
  number={5},
  pages={1--12},
  year={2019}
}

@article{thomas2018tensor,
  title={Tensor field networks: Rotation- and translation-equivariant neural networks for {3D} point clouds},
  author={Thomas, Nathaniel and Smidt, Tess and Kearnes, Steven and Yang, Lusann and Li, Li and Kohlhoff, Kai and Riley, Patrick},
  journal={arXiv preprint arXiv:1802.08219},
  year={2018}
}

@inproceedings{fuchs2020se3,
  title={{SE}(3)-transformers: {3D} roto-translation equivariant attention networks},
  author={Fuchs, Fabian and Worrall, Daniel and Fischer, Volker and Welling, Max},
  booktitle={Advances in Neural Information Processing Systems (NeurIPS)},
  year={2020}
}

@inproceedings{weiler20183d,
  title={{3D} steerable {CNN}s: Learning rotationally equivariant features in volumetric data},
  author={Weiler, Maurice and Geiger, Mario and Welling, Max and Boomsma, Wouter and Cohen, Taco},
  booktitle={Advances in Neural Information Processing Systems (NeurIPS)},
  year={2018}
}

@inproceedings{cohen2016group,
  title={Group equivariant convolutional networks},
  author={Cohen, Taco and Welling, Max},
  booktitle={International Conference on Machine Learning (ICML)},
  year={2016}
}

@article{boscain2014hypoelliptic,
  title={Hypoelliptic diffusion and human vision: A semidiscrete new twist},
  author={Boscain, Ugo and Chertovskih, Roman A and Gauthier, Jean-Paul and Remizov, Alexey O},
  journal={SIAM Journal on Imaging Sciences},
  volume={7},
  number={2},
  pages={669--695},
  year={2014}
}

@inproceedings{bekkers2018roto,
  title={Roto-translation covariant convolutional networks for medical image analysis},
  author={Bekkers, Erik J and Lafarge, Maxime W and Veta, Mitko and Eppenhof, Koen AJ and Pluim, Josien PW and Duits, Remco},
  booktitle={Medical Image Computing and Computer Assisted Intervention (MICCAI)},
  year={2018}
}

@inproceedings{bekkers2020bspline,
  title={B-spline {CNN}s on {L}ie groups},
  author={Bekkers, Erik J},
  booktitle={International Conference on Learning Representations (ICLR)},
  year={2020}
}

@inproceedings{sun2009concise,
  title={A concise and provably informative multi-scale signature based on heat diffusion},
  author={Sun, Jian and Ovsjanikov, Maks and Guibas, Leonidas},
  booktitle={Computer Graphics Forum (Proc. SGP)},
  volume={28},
  pages={1383--1392},
  year={2009}
}

@article{crane2013geodesics,
  title={Geodesics in heat: A new approach to computing distance based on heat flow},
  author={Crane, Keenan and Weischedel, Clarisse and Wardetzky, Max},
  journal={ACM Transactions on Graphics},
  volume={32},
  number={5},
  pages={1--11},
  year={2013}
}

@inproceedings{monti2017geometric,
  title={Geometric deep learning on graphs and manifolds using mixture model {CNN}s},
  author={Monti, Federico and Boscaini, Davide and Masci, Jonathan and Rodol{\`a}, Emanuele and Svoboda, Jan and Bronstein, Michael M},
  booktitle={IEEE Conference on Computer Vision and Pattern Recognition (CVPR)},
  year={2017}
}

@inproceedings{fleishman2003bilateral,
  title={Bilateral mesh denoising},
  author={Fleishman, Shachar and Drori, Iddo and Cohen-Or, Daniel},
  booktitle={ACM SIGGRAPH},
  pages={950--953},
  year={2003}
}

@inproceedings{guerrero2018pcpnet,
  title={{PCPN}et: Learning local shape properties from raw point clouds},
  author={Guerrero, Paul and Kleiman, Yanir and Ovsjanikov, Maks and Mitra, Niloy J},
  booktitle={Computer Graphics Forum (Proc. Eurographics)},
  volume={37},
  pages={75--85},
  year={2018}
}

@article{mirebeau2018fast,
  title={Fast-marching methods for curvature penalized shortest paths},
  author={Mirebeau, Jean-Marie},
  journal={Journal of Mathematical Imaging and Vision},
  volume={60},
  number={6},
  pages={784--815},
  year={2018}
}

@article{bekkers2015pde,
  title={A {PDE} approach to data-driven sub-{R}iemannian geodesics in {$SE(2)$}},
  author={Bekkers, Erik J and Duits, Remco and Mashtakov, Alexey and Sanguinetti, Gonzalo R},
  journal={SIAM Journal on Imaging Sciences},
  volume={8},
  number={4},
  pages={2740--2770},
  year={2015}
}

@article{rashevskii1938connectability,
  title={About connecting two points of a completely nonholonomic space by an admissible curve},
  author={Rashevskii, Petr K},
  journal={Uchenye Zapiski Pedagogicheskogo Instituta imeni Libknechta},
  volume={2},
  pages={83--94},
  year={1938}
}

@article{taubin1995signal,
  title={A signal processing approach to fair surface design},
  author={Taubin, Gabriel},
  journal={ACM SIGGRAPH},
  pages={351--358},
  year={1995}
}

@article{bekkers2018nilpotent,
author = {Bekkers, Erik J. and Chen, Da and Portegies, Jorg M.},
title = {Nilpotent Approximations of Sub-Riemannian Distances for Fast Perceptual Grouping of Blood Vessels in 2D and 3D},
year = {2018},
issue_date = {July      2018},
publisher = {Kluwer Academic Publishers},
address = {USA},
volume = {60},
number = {6},
issn = {0924-9907},
url = {https://doi.org/10.1007/s10851-018-0787-z},
doi = {10.1007/s10851-018-0787-z},
journal = {J. Math. Imaging Vis.},
month = jul,
pages = {882–899},
numpages = {18}
}

@inproceedings{kessler2026sinkhorn,
  title     = {Sinkhorn Normalization of Diffusion Kernels},
  author    = {Kessler, Nathan and Magnet, Robin and Feydy, Jean},
  booktitle = {International Conference on Machine Learning (ICML)},
  year      = {2026},
}

@article{schunck2021pheno4d,
  title={Pheno4D: A spatio-temporal dataset of maize and tomato plant point clouds for phenotyping and advanced plant analysis},
  author={Schunck, David and Magistri, Federico and Rosu, Radu Alexandru and Cornelißen, André and Chebrolu, Nived and Paulus, Stefan and Léon, Jens and Behnke, Sven and Stachniss, Cyrill and Kuhlmann, Heiner and Klingbeil, Lasse},
  journal={PLOS ONE},
  volume={16},
  number={8},
  pages={e0256340},
  year={2021},
  publisher={Public Library of Science},
  doi={10.1371/journal.pone.0256340}
}

@article{Mitchell1985OnCM,
  title={On Carnot-Carath{\'e}odory metrics},
  author={John Leslie Mitchell},
  journal={Journal of Differential Geometry},
  year={1985},
  volume={21},
  pages={35-45},
  url={https://api.semanticscholar.org/CorpusID:120608865}
}

@inproceedings{Bogo2014faust,
      title = {{FAUST}: Dataset and evaluation for {3D} mesh registration},
      author = {Bogo, Federica and Romero, Javier and Loper, Matthew and Black, Michael J.},
      booktitle = {Proceedings IEEE Conf. on Computer Vision and Pattern Recognition (CVPR)},
      address = {Piscataway, NJ, USA},
      publisher = {IEEE},
      month = jun,
      year = {2014}
    }

@inproceedings{Rusinkiewicz2001ptp,
  author = {Szymon Rusinkiewicz and Marc Levoy},
  title = {Efficient Variants of the {ICP} Algorithm},
  booktitle = {Third International Conference on 3D Digital Imaging and Modeling (3DIM)},
  pages = {145--152},
  year = {2001},
  organization = {IEEE}
}

@Article{kerbl3Dgaussians,
      author       = {Kerbl, Bernhard and Kopanas, Georgios and Leimk{\"u}hler, Thomas and Drettakis, George},
      title        = {3D Gaussian Splatting for Real-Time Radiance Field Rendering},
      journal      = {ACM Transactions on Graphics},
      number       = {4},
      volume       = {42},
      month        = {July},
      year         = {2023},
      url          = {https://repo-sam.inria.fr/fungraph/3d-gaussian-splatting/}
}

@article{charon2013varifold,
author = {Charon, Nicolas and Trouv{\'e}, Alain},
title = {The Varifold Representation of Nonoriented Shapes for Diffeomorphic Registration},
journal = {SIAM Journal on Imaging Sciences},
volume = {6},
number = {4},
pages = {2547-2580},
year = {2013},
doi = {10.1137/130918885},
URL = {https://doi.org/10.1137/130918885},
eprint = {https://doi.org/10.1137/130918885}
}

@inproceedings{kaltenmark2017varifold,
author = {Kaltenmark, Irène and Charlier, Benjamin and Charon, Nicolas},
year = {2017},
month = {07},
journal = {Proceedings of the IEEE/CVF Conference on Computer Vision and Pattern Recognition (CVPR)},
title = {A General Framework for Curve and Surface Comparison and Registration with Oriented Varifolds},
doi = {10.1109/CVPR.2017.487}
}

@article{buet2016varifold,
author = {Buet, Blanche and Leonardi, Gian and Masnou, Simon},
year = {2017},
month = {11},
pages = {},
title = {A Varifold Approach to Surface Approximation},
volume = {226},
journal = {Archive for Rational Mechanics and Analysis},
doi = {10.1007/s00205-017-1141-0}
}

@Inbook{bellaiche1996,
author="Bella{\"i}che, Andr{\'e}",
editor="Bella{\"i}che, Andr{\'e}
and Risler, Jean-Jacques",
title="The tangent space in sub-Riemannian geometry",
bookTitle="Sub-Riemannian Geometry",
year="1996",
publisher="Birkh{\"a}user Basel",
address="Basel",
pages="1--78",
isbn="978-3-0348-9210-0",
doi="10.1007/978-3-0348-9210-0_1",
url="https://doi.org/10.1007/978-3-0348-9210-0_1"
}

@article{Sharp2020ALF,
  title={A Laplacian for Nonmanifold Triangle Meshes},
  author={Nicholas Sharp and Keenan Crane},
  journal={Computer Graphics Forum},
  year={2020},
  volume={39},
  url={https://api.semanticscholar.org/CorpusID:221607415}
}

@article{Petrov2026blowup,
author = {Petrov, Alice and Nabizadeh, Mohammad Sina and Dodik, Ana and Solomon, Justin},
title = {Tangent Blow-Ups for Processing Non-Manifold Geometry},
journal = {Computer Graphics Forum},
year ={2026},
doi = {https://doi.org/10.1111/cgf.70527},
url = {https://onlinelibrary.wiley.com/doi/abs/10.1111/cgf.70527},
eprint = {https://onlinelibrary.wiley.com/doi/pdf/10.1111/cgf.70527}
}

@misc{gomez2026blobsspokeshighfidelitysurface,
    title={From Blobs to Spokes: High-Fidelity Surface Reconstruction via Oriented Gaussians},
    author={Gomez, Diego and Gu{\'e}don, Antoine and Maruani, Nissim and Gong, Bingchen and Ovsjanikov, Maks},
    year={2026},
    journal = {Proceedings of the European Conference on Computer Vision (ECCV)},
}

@inproceedings{Bellaard2025RotoTranslationIM,
  title={Roto-Translation Invariant Metrics on Position-Orientation Space},
  author={Gijs Bellaard and Bart M. N. Smets},
  booktitle={International Conference on Geometric Science of Information},
  year={2025},
  url={https://api.semanticscholar.org/CorpusID:277596215}
}

@inproceedings{Portegies2015NewAO,
  title={New Approximation of a Scale Space Kernel on SE(3) and Applications in Neuroimaging},
  author={Jorg M. Portegies and Gonzalo Sanguinetti and Stephan P. L. Meesters and Remco Duits},
  booktitle={Scale Space and Variational Methods in Computer Vision},
  year={2015},
  url={https://api.semanticscholar.org/CorpusID:7968693}
}

@inproceedings{thomas2019kpconv,
  author    = {Thomas, Hugues and Qi, Charles R. and Deschaud, Jean-Emmanuel and Marcotegui, Beatriz and Goulette, Fran\c{c}ois and Guibas, Leonidas J.},
  title     = {KPConv: Flexible and Deformable Convolution for Point Clouds},
  booktitle = {Proceedings of the IEEE/CVF International Conference on Computer Vision (ICCV)},
  year      = {2019},
  pages     = {6410--6419}
}

@inproceedings{wu2024ptv3,
  title={Point Transformer V3: Simpler, Faster, Stronger},
  author={Wu, Xiaoyang and Jiang, Li and Wang, Peng-Shuai and Liu, Zhijian and Liu, Xihui and Qiao, Yu and Ouyang, Wanli and He, Tong and Zhao, Hengshuang},
  booktitle={Proceedings of the IEEE/CVF Conference on Computer Vision and Pattern Recognition (CVPR)},
  year={2024},
  pages={4840-4851}
}

@article{Guennebaud2007algebraic,
author = {Guennebaud, Ga{\"e}l and Gross, Markus},
title = {Algebraic point set surfaces},
year = {2007},
issue_date = {July 2007},
publisher = {Association for Computing Machinery},
address = {New York, NY, USA},
volume = {26},
number = {3},
issn = {0730-0301},
url = {https://doi.org/10.1145/1276377.1276406},
doi = {10.1145/1276377.1276406},
journal = {ACM Trans. Graph.},
month = jul,
pages = {23–es},
numpages = {10}
}

@article{Öztireli2009feat,
author = {Öztireli, A. C. and Guennebaud, G. and Gross, M.},
title = {Feature Preserving Point Set Surfaces based on Non-Linear Kernel Regression},
journal = {Computer Graphics Forum},
volume = {28},
number = {2},
pages = {493-501},
doi = {https://doi.org/10.1111/j.1467-8659.2009.01388.x},
url = {https://onlinelibrary.wiley.com/doi/abs/10.1111/j.1467-8659.2009.01388.x},
eprint = {https://onlinelibrary.wiley.com/doi/pdf/10.1111/j.1467-8659.2009.01388.x},
year = {2009}
}

@book{geiges2008introduction,
  title     = {An Introduction to Contact Topology},
  author    = {Geiges, Hansj{\"o}rg},
  series    = {Cambridge Studies in Advanced Mathematics},
  volume    = {109},
  year      = {2008},
  publisher = {Cambridge University Press},
  address   = {Cambridge},
  isbn      = {9780521865852}
}

@article{federer1959curvature,
  author  = {Federer, Herbert},
  title   = {Curvature measures},
  journal = {Transactions of the American Mathematical Society},
  volume  = {93},
  number  = {3},
  pages   = {418--491},
  year    = {1959}
}

@article{boissonnat2019reach,
  author  = {Boissonnat, Jean-Daniel and Lieutier, Andr{\'e} and Wintraecken, Mathijs},
  title   = {The reach, metric distortion, geodesic convexity and the variation of tangent spaces},
  journal = {Journal of Applied and Computational Topology},
  volume  = {3},
  number  = {1},
  pages   = {29--58},
  year    = {2019}
}
